%% file: main.tex
\documentclass[10pt,twocolumn,letterpaper]{article}

\usepackage[pagenumbers]{wacv} 

\input{preamble}

\definecolor{wacvblue}{rgb}{0.21,0.49,0.74}
\usepackage[pagebackref,breaklinks,colorlinks,allcolors=wacvblue]{hyperref}

\def\wacvPaperID{1441} 
\def\confName{WACV}
\def\confYear{2027}

\title{\ours: Zero-Shot Object Out-of-Distribution Detection\\ by Object Class Prompting of the \texttt{SAM3-Image} Model}

\author{Lucas Görnhardt \quad Timo Bartels \quad Tim Fingscheidt\\
Technische Universität Braunschweig,  Institute for Communications Technology\\
{\tt\small \{lucas.goernhardt, timo.bartels, t.fingscheidt\}@tu-bs.de}
}

\begin{document}
\maketitle
\input{sec/0_abstract}

\input{sec/1_intro}
\input{sec/02_related_works}
\input{sec/03_methods}
\input{sec/04_experimental_setup}
\input{sec/05_experimental_results}
\input{sec/06_conclusions}
 \section*{Acknowledgments}
 The research leading to these results is funded by the German Federal Ministry for Economic Affairs and Energy within the project “Safe AI Engineering – Sicherheitsargumentation befähigendes AI Engineering \"uber den gesamten Lebenszyklus einer KI-Funktion". The authors would like to thank the consortium for the successful cooperation.
{
    \small
    \bibliographystyle{ieeenat_fullname}
    \bibliography{object_detection}
}
\clearpage
\appendix
\input{app/sec/ap_results}
\input{app/sec/full_test_set}
\input{app/sec/abl_iou}
\input{app/sec/openworld_setting}
\input{app/sec/computational_complexity}
\input{app/sec/few_shot_methods}
\input{app/sec/localization_error}
\input{app/sec/qualitative_results}

\end{document}

%% file: preamble.tex
\usepackage{pifont}
\newcommand{\cmark}{\ding{51}}%
\newcommand{\xmark}{\ding{55}}%
\usepackage{makecell}
\usepackage{multirow}
\usepackage{tikz}
\usepackage{pgfplots}
\pgfplotsset{compat=1.18}
\usepackage{xspace}
\usetikzlibrary{
  backgrounds,
  calc,
  positioning,
  quotes,
  patterns,
  spy,
  arrows.meta,
  decorations.pathmorphing,
  shapes.geometric,
  fit,
  decorations.pathreplacing,
  matrix
}
\usepackage{siunitx}

\InputIfFileExists{tubs_colors}{}{}

\newcommand{\ours}{\texttt{SAM3-O2D2}\xspace}

%% file: tubs_colors.tex
\definecolor{tu1}{rgb}{1.0000, 0.8039, 0.0000}
\definecolor{tu11}{rgb}{1.0000, 0.8627, 0.3020}
\definecolor{tu12}{rgb}{1.0000, 0.9020, 0.4980}
\definecolor{tu13}{rgb}{1.0000, 0.9412, 0.6980}
\definecolor{tu14}{rgb}{1.0000, 0.9608, 0.8000}

\definecolor{tu2}{rgb}{0.9804, 0.4314, 0.0000}
\definecolor{tu21}{rgb}{0.9882, 0.6039, 0.3020}
\definecolor{tu22}{rgb}{0.9882, 0.7137, 0.4980}
\definecolor{tu23}{rgb}{0.9922, 0.8275, 0.6980}
\definecolor{tu24}{rgb}{0.9961, 0.8863, 0.8000}

\definecolor{tu3}{rgb}{0.6902, 0.0000, 0.2745}
\definecolor{tu31}{rgb}{0.7529, 0.2000, 0.4196}
\definecolor{tu32}{rgb}{0.8431, 0.4980, 0.6353}
\definecolor{tu33}{rgb}{0.9216, 0.7490, 0.8196}
\definecolor{tu34}{rgb}{0.9529, 0.8510, 0.8902}

\definecolor{tu4}{rgb}{0.4863, 0.8039, 0.9020}
\definecolor{tu41}{rgb}{0.6431, 0.8627, 0.9333}
\definecolor{tu42}{rgb}{0.7412, 0.9020, 0.9490}
\definecolor{tu43}{rgb}{0.8431, 0.9412, 0.9686}
\definecolor{tu44}{rgb}{0.8980, 0.9608, 0.9804}

\definecolor{tu5}{rgb}{0.0000, 0.5020, 0.7059}
\definecolor{tu51}{rgb}{0.3020, 0.6510, 0.7961}
\definecolor{tu52}{rgb}{0.5490, 0.7765, 0.8667}
\definecolor{tu53}{rgb}{0.7490, 0.8745, 0.9255}
\definecolor{tu54}{rgb}{0.8510, 0.9255, 0.9569}

\definecolor{tu6}{rgb}{0.0000, 0.3255, 0.4549}
\definecolor{tu61}{rgb}{0.2510, 0.4941, 0.5922}
\definecolor{tu62}{rgb}{0.5490, 0.6941, 0.7529}
\definecolor{tu63}{rgb}{0.7490, 0.8314, 0.8627}
\definecolor{tu64}{rgb}{0.8510, 0.8980, 0.9176}

\definecolor{tu7}{rgb}{0.0314, 0.0314, 0.0314}
\definecolor{tu71}{rgb}{0.3725, 0.3725, 0.3725}
\definecolor{tu72}{rgb}{0.5882, 0.5882, 0.5882}
\definecolor{tu73}{rgb}{0.7529, 0.7529, 0.7529}
\definecolor{tu74}{rgb}{0.8667, 0.8667, 0.8667}

\definecolor{tu8}{rgb}{0.7765, 0.9333, 0.0000}
\definecolor{tu81}{rgb}{0.8431, 0.9529, 0.3020}
\definecolor{tu82}{rgb}{0.8863, 0.9647, 0.4980}
\definecolor{tu83}{rgb}{0.9333, 0.9804, 0.6980}
\definecolor{tu84}{rgb}{0.9569, 0.9882, 0.8000}

\definecolor{tu9}{rgb}{0.5373, 0.6431, 0.0000}
\definecolor{tu91}{rgb}{0.6784, 0.7490, 0.3020}
\definecolor{tu92}{rgb}{0.7686, 0.8196, 0.4980}
\definecolor{tu93}{rgb}{0.8588, 0.8941, 0.6980}
\definecolor{tu94}{rgb}{0.9059, 0.9294, 0.8000}

\definecolor{tu10}{rgb}{0.0000, 0.4431, 0.3373}
\definecolor{tu101}{rgb}{0.3020, 0.6118, 0.5373}
\definecolor{tu102}{rgb}{0.5490, 0.7490, 0.7020}
\definecolor{tu103}{rgb}{0.7490, 0.8588, 0.8353}
\definecolor{tu104}{rgb}{0.8549, 0.9176, 0.9059}

\definecolor{tu110}{rgb}{0.8000, 0.0000, 0.6000}
\definecolor{tu111}{rgb}{0.8706, 0.3490, 0.7412}
\definecolor{tu112}{rgb}{0.9216, 0.6000, 0.8392}
\definecolor{tu113}{rgb}{0.9608, 0.8000, 0.9216}
\definecolor{tu114}{rgb}{0.9804, 0.8980, 0.9608}

\definecolor{tu120}{rgb}{0.4627, 0.0000, 0.4627}
\definecolor{tu121}{rgb}{0.5961, 0.2510, 0.5961}
\definecolor{tu122}{rgb}{0.7294, 0.4980, 0.7294}
\definecolor{tu123}{rgb}{0.8392, 0.6980, 0.8392}
\definecolor{tu124}{rgb}{0.9216, 0.8510, 0.9216}

\definecolor{tu130}{rgb}{0.4627, 0.0000, 0.3294}
\definecolor{tu131}{rgb}{0.6118, 0.3020, 0.5333}
\definecolor{tu132}{rgb}{0.7569, 0.5490, 0.6980}
\definecolor{tu133}{rgb}{0.8667, 0.7490, 0.8314}
\definecolor{tu134}{rgb}{0.9216, 0.8510, 0.9020}

\definecolor{tu140}{rgb}{0.7019, 1, 0.7019}
\definecolor{tu141}{rgb}{0.7216, 1, 0.713}
\definecolor{tu142}{rgb}{0.8039, 1, 0.7725}
\definecolor{tu143}{rgb}{0.9098, 0.9921, 0.8470}

\definecolor{tu151}{rgb}{0.1607, 0.1960, 0.1529}
\definecolor{tu152}{rgb}{0.4549, 0.4941, 0.4235}
\definecolor{tu153}{rgb}{0.70196, 0.6941, 0.6392}

%% file: sec/0_abstract.tex
\begin{abstract}
Object detectors have shown remarkable performance in various fields, among these medical imaging, surveillance, and autonomous driving. However, they are prone to overconfidence when encountering unseen objects in real-world deployments, causing potential safety issues. To address this, detecting out-of-distribution (OOD) objects is essential for reliable object detection. Modern approaches leverage the broad semantic knowledge of foundation models such as \texttt{CLIP} for post-hoc few- and zero-shot OOD detection. However, these methods typically perform OOD assessment in feature space, which can be sensitive to object detector localization errors and variations in object appearance.
Moreover, the current state-of-the-art (SOTA) zero-shot method performs computationally costly diffusion in inference.
In this work, for our proposed zero-shot object OOD detection method \ours, we employ the \texttt{SAM3-image} foundation model in an efficient manner. Specifically, we prompt \texttt{SAM3} only with the object detector’s predicted classes and compare the predictions of the object detector and \texttt{SAM3}. An object is in-distribution (ID), if \texttt{SAM3} also detects an object at the corresponding location. If  \texttt{SAM3} does not detect the prompted object, this indicates a mismatch between the detector’s prediction and the image content, suggesting that the object is OOD. Experimental results show that our method significantly surpasses the so-far zero-shot SOTA method. Specifically, we achieve new SOTA AuROC and FPR95 metrics over both ID datasets Pascal-VOC and BDD100K and both OOD datasets MS-COCO and OpenImages. 

\end{abstract}

%% file: sec/1_intro.tex
\section{Introduction}
\label{sec:intro}
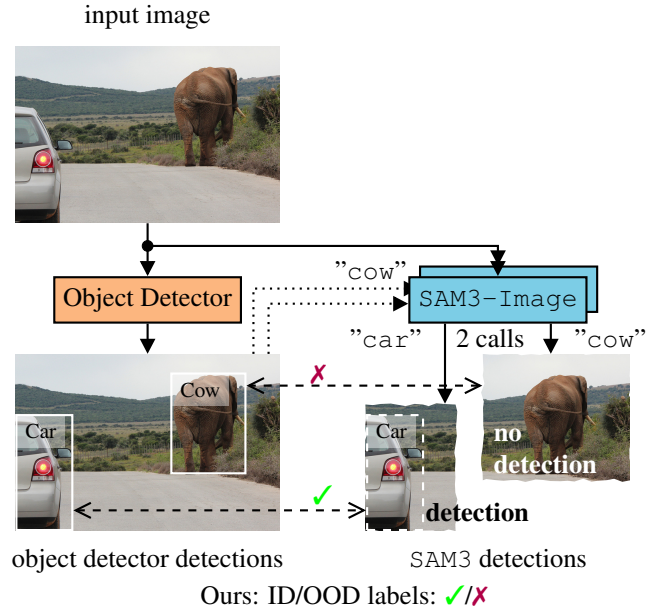
\begin{figure}[t!]
    \centering
    \input{figures/method_preview}
    \caption{In \textbf{our zero-shot object OOD detection}, we prompt \texttt{SAM3-image} \cite{carion2026sam3segmentconcepts} with objects from the object detector for object OOD detection.}
\label{fig:method_highlevel}
\end{figure}

Detecting out-of-distribution (OOD) objects is crucial for the safe deployment of object detection systems in the real world.
Object detectors are typically trained on a closed set of in-distribution (ID) classes, resulting in overconfidence when encountering unseen objects during deployment.
Motivated by this challenge, extensive research has studied image-level OOD detection \cite{Miyai2025, Wang_2023_ICCV, Liu2024, ding2024zero, Ming2022}, where the entire image is treated as a single sample and is classified as either ID or OOD.
However, this setting is insufficient for many real-world applications, such as autonomous driving \cite{Fingscheidt2022}, where images contain multiple objects that require independent OOD assessment.

Consequently, object OOD detection has emerged as a more suitable formulation for such object-level requirements \cite{Du2022, Xuefeng2022b}.
In this task, each detected object must be independently classified as ID (\textcolor{green}{\cmark}) or OOD (\textcolor{tu3}{\xmark}), see \cref{fig:method_highlevel}.
Classical approaches \cite{Du2022, Xuefeng2022b} rely on training of additional detection modules alongside the object detector to distinguish between ID and OOD objects.
However, these methods require access to full ID training data and need to be retrained for each newly deployed object detector.
Moreover, the performance of such small models falls behind the performance of modern vision-language foundation models such as \texttt{CLIP} \cite{radford2021learningtransferablevisualmodels}.

Recent few-shot methods \cite{zhang2025runaobjectleveloutofdistributiondetection, Li_2026_CVPR} address these limitations by leveraging \texttt{CLIP} \cite{radford2021learningtransferablevisualmodels} for object OOD detection.
They use pre-trained image and text encoders to extract local and global visual representations and compare them with text embeddings obtained from class prompts.
While these approaches achieve strong performance, they still require fine-tuning of additional modules for each ID dataset.
In contrast, \texttt{RONIN} \cite{Nguyen2026} proposes a zero-shot approach based on class-conditioned inpainting.
Given a predicted object and its class label, \texttt{RONIN} uses a text-conditioned diffusion model to inpaint the object region according to the predicted class and then measures the agreement between the original object, the inpainted object, and the text prompt using vision-language feature similarities.
Despite achieving state-of-the-art (SOTA) zero-shot object OOD detection performance, \texttt{RONIN} has two important limitations.
First, its reliance on diffusion-based inpainting leads to high computational cost, which limits its practical applicability.
Second, its OOD decision is based on similarities in an abstract feature space, which is not necessarily robust and focused, as the embeddings entangle object appearance such as size and lighting and are not robust against detector localization errors, i.e., poorly localized bounding boxes.

We, however, take a different approach. In contrast to the \texttt{CLIP}-based methods \cite{zhang2025runaobjectleveloutofdistributiondetection, Li_2026_CVPR, Nguyen2026} we leverage a different foundation model, namely \texttt{SAM3-image} \cite{carion2026sam3segmentconcepts} and propose \ours. In contrast to \texttt{CLIP}- and diffusion-based approaches \cite{zhang2025runaobjectleveloutofdistributiondetection, Li_2026_CVPR, Nguyen2026}, our method performs OOD detection directly at the prediction level.
It achieves SOTA zero-shot object OOD detection performance while requiring lower computational cost than diffusion-based inference.
The high-level concept is illustrated in \cref{fig:method_highlevel}.
The input image is first processed by some object detector, which produces object predictions such as \texttt{car} and \texttt{cow}.
The predicted class labels are used as text prompts for two \texttt{SAM3-image} calls, which returns objects corresponding to each prompted concept.
As illustrated in \cref{fig:method_highlevel}, \texttt{SAM3-image} confirms the detector's \texttt{car} prediction at the same location, but does not detect a \texttt{cow}.
This mismatch indicates that the visual content is inconsistent with the predicted class \texttt{cow} and instead corresponds to an OOD object, here an \texttt{elephant}.
Thus, our method marks predictions confirmed by \texttt{SAM3-image} as ID (\textcolor{green}{\cmark}) and unmatched predictions as OOD (\textcolor{tu3}{\xmark}), leveraging its broad semantic knowledge.
Moreover, by operating directly in the detection space rather than in an abstract feature space, our method provides a more interpretable OOD decision.
This simplifies failure-case analysis, which is important for practical deployment, and improves robustness to near-OOD objects, as reflected by our empirical results.

Our contributions are as follows.
First, we introduce a framework for leveraging off-the-shelf text-promptable foundation detection models, such as \texttt{SAM3}, for object OOD detection and propose \ours, which constantly improves with the rapid developments in the field of foundation models over time.
Second, our proposed \ours requires no retraining and can be paired with arbitrary object detectors. It is straightforward to implement, making it highly suitable for practical deployment.
Third, we evaluate our method on two ID datasets and two OOD datasets covering a diverse range of semantic scenes, and finally on one open-world setup, showing that it substantially outperforms the current SOTA zero-shot object OOD detection method \texttt{RONIN} \cite{Nguyen2026}, while even requiring lower computational complexity.

The remainder of the paper is structured as follows. \Cref{sec:rel_works} summarizes the related works. \Cref{sec:methods} details the proposed method for zero-shot object OOD detection. \Cref{sec:exp_setup} describes the experimental setup, including datasets and metrics. \Cref{sec:results} presents the experimental results. Finally, we conclude in \Cref{sec:conclusions}.

%% file: figures/method_preview.tex
\pgfsetlayers{background,main}
\begin{tikzpicture}

\node[inner sep=0, minimum width=1.524cm, minimum height=0.752cm] (input_image) {\includegraphics[width=0.2\textwidth]{figures/images/000000166664.jpg}};

\node[align=center, rectangle, draw, below = 0.7 of input_image, fill=tu22, minimum width=1.5cm, minimum height=0.6cm, thick] (od) {Object Detector};

\node[inner sep=0, minimum width=1.524cm, minimum height=0.752cm, below = 0.4cm of od] (od_image) {\includegraphics[width=0.2\textwidth]{figures/images/000000166664.jpg}};

\node[align=center, rectangle, draw, right = 2.2cm of od, fill=tu4, minimum width=1.50cm, minimum height=0.6cm, thick] (sam3) {\texttt{SAM3-Image}};

\begin{pgfonlayer}{background}
\node[align=center, rectangle, draw, above right = 0.2cm of sam3.center, anchor=center, fill=tu4, minimum width=1.50cm, minimum height=0.6cm, thick] (sam3_2) {\texttt{SAM3-Image}};
\end{pgfonlayer}


\node[rectangle, draw=white, minimum width=0.74cm, minimum height=1.52cm, thick, anchor=west] at (-1.75,-4.47cm) (bbox1) {};
\node[rectangle,fill=white, fill opacity=0.4, below = 0.0 of bbox1.north, thick, anchor=north, xshift=-1.3, text=black, text opacity=1] (bbox1_tag) {\footnotesize{Car}};

\node[rectangle, draw=white, minimum width=0.74cm, minimum height=1.52cm, thick, anchor=west] at (2.89cm,-4.47cm) (bbox2) {};
\node[rectangle, fill=white, fill opacity=0.4, below = 0.0 of bbox2.north, thick, anchor=north, xshift=-1.3, text=black, text opacity=1] (bbox2_tag) {\footnotesize{Car}};

\node[rectangle, draw=white, minimum width=0.98cm, minimum height=1.32cm, thick, anchor=west] at (0.30cm,-3.82cm) (bbox3) {};
\node[rectangle,fill=white, fill opacity=0.4, above = 0.0 of bbox3.north, thick, anchor=north, xshift=-3.0, text=black, text opacity=1] (bbox3_tag) {\footnotesize{Cow}};

\def\imgfile{figures/images/000000166664.jpg}

\node[inner sep=0, minimum width=1.524cm, minimum height=0.752cm,
      below = 0.4cm of sam3] (sam3_image)
      {\includegraphics[width=0.2\textwidth]{\imgfile}};

\coordinate (blA) at ($(bbox2.west) + (0.0cm, 0.9cm)$);   
\coordinate (blB) at ($(bbox2.south) + (0.8cm, 0.0cm)$);  

\coordinate (trA) at ($(sam3_image.north) + (-0.2cm, 0.0cm)$); 
\coordinate (trB) at ($(sam3_image.east)  + ( 0.0cm,-0.5cm)$); 

\fill[white] (sam3_image.south west) rectangle (sam3_image.north east);

\coordinate (blA) at ($(bbox2.west) + (0.0cm, 0.9cm)$);   
\coordinate (blB) at ($(bbox2.south) + (0.8cm, 0.0cm)$);  

\coordinate (blLeft)   at (sam3_image.west |- blA);   
\coordinate (blBottom) at (sam3_image.south -| blB);  

\begin{scope}
  \clip[
    decorate,
    decoration={random steps, segment length=3pt, amplitude=0.7pt}
  ]
    (blLeft) -- (blA) -| (blB) -- (blBottom) -- (sam3_image.south west) -- cycle;

  \node[inner sep=0, anchor=center] at (sam3_image.center)
    {\includegraphics[width=0.2\textwidth]{\imgfile}};
\end{scope}

\begin{scope}
  \clip[
    decorate,
    decoration={random steps, segment length=3pt, amplitude=0.7pt}
  ]
    (trA) -- (sam3_image.north east) -- (trB) -| (trA);

  \node[inner sep=0, anchor=center] at (sam3_image.center)
    {\includegraphics[width=0.2\textwidth]{\imgfile}};
\end{scope}

\draw[thick,-{Triangle[length=2mm,width=2mm]}] (input_image) -- (od);

\coordinate (split3) at ($(input_image.south) + (0.0cm, -0.3cm)$);
\filldraw (split3) circle (2pt);  

\draw[thick,-{Triangle[length=2mm,width=2mm]}] (split3) -| (sam3);
\draw[thick,-{Triangle[length=2mm,width=2mm]}] ($(sam3_2.north) + (-0.14cm, 0.1cm)$) -| ($(sam3_2.north) + (-0.14cm, 0.0cm)$);

\draw[thick,-{Triangle[length=2mm,width=2mm]}] (od) -- (od_image);
\begin{pgfonlayer}{background}
\draw[thick,-{Triangle[length=2mm,width=2mm]}, dotted] ($(od_image.north) + (1.4cm, 0.0cm)$) |- ($(sam3.west) + (0.14,0.15cm)$);

\draw[thick,-{Triangle[length=2mm,width=2mm]}, dotted] ($(od_image.north) + (1.6cm, 0.0cm)$) |- ($(sam3.west) + (0.0,-0.05cm)$);
\end{pgfonlayer}

\draw[thick,-{Triangle[length=2mm,width=2mm]}] ($(sam3.south) + (0.7cm, 0.0cm)$) -- ($(sam3_image.north) + (0.7cm, 0.0cm)$);

\node [above = 0.1cm of input_image.north, anchor = south, align=center, thick] (input_image_desc)  {input image};

\node [below = 0.1cm of sam3_image.south, anchor = north, align=center, thick] (input_image_desc)  {\texttt{SAM3} detections};

\node [below = 0.1cm of od_image.south, anchor = north, align=center, thick] (input_image_desc)  {object detector detections};

\node [right = -0.3cm of sam3.east, anchor = west, align=center, thick, yshift=-0.50cm] (labels)  {"\texttt{cow}"};

\node [right = 1.3cm of od.east, anchor = west, align=center, thick, yshift=-0.50cm] (labels)  {"\texttt{car}"};

\node [right = 1.1cm of od.east, anchor = west, align=center, thick, yshift=0.40cm] (labels_2)  {"\texttt{cow}"};

\node [right = 2.7cm of od.east, anchor = west, align=center, thick, yshift=-0.5cm] (calls)  {2 calls};

\node [right = 3.2cm of od.east, anchor = west, align=left, thick, yshift=-2.0cm, text=white] (nd)  {\textbf{no} \\ \textbf{detection}};

\node [right = 2.3cm of od.east, anchor = west, align=left, thick, yshift=-2.8cm] (nd)  {\textbf{detection}};

\node[rectangle, draw=white, minimum width=0.74cm, minimum height=1.52cm, thick, anchor=west, dashed] at (2.89cm,-4.47cm) (bbox2) {};
\node[rectangle, fill=white, fill opacity=0.4, below = 0.0 of bbox2.north, thick, anchor=north, xshift=-1.3, text=black, text opacity=1] (bbox2_tag) {\footnotesize{Car}};

\draw[
  thick,
  {Straight Barb[length=2mm,width=2mm]}-{Straight Barb[length=2mm,width=2mm]},
  dashed
] ($(bbox1.east) + (0,-0.5cm)$) -- ($(bbox2.west) + (0,-0.5cm)$);

\draw[
  thick,
  {Straight Barb[length=2mm,width=2mm]}-{Straight Barb[length=2mm,width=2mm]},
  dashed
] ($(bbox3.east) + (3.12cm,+0.5cm)$) -- ($(bbox3.east) + (0cm,+0.5cm)$);

\node [right = 0.95cm of bbox3.east, anchor = center, align=center, thick, yshift=0.65cm] (cross)  {\textcolor{tu3}{\xmark}};

\node [right = 3.3cm of bbox1.east, anchor = center, align=center, thick, yshift=-0.30cm] (mark)  {\textcolor{green}{\cmark}};


\node [below = -0.1cm of input_image_desc.south, anchor = north, align=center, thick, xshift=2.6cm] (input_image_desc_baldwiederweg)  {Ours: ID/OOD labels: \textcolor{green}{\cmark}/\textcolor{tu3}{\xmark}};



\draw[thick,-{Triangle[length=2mm,width=2mm]}] ($(sam3.south) + (-0.7cm, 0.0cm)$) -- ($(sam3.south) + (-0.7cm, -1.06cm)$);

\end{tikzpicture}

%% file: sec/02_related_works.tex
\section{Related Works}
\label{sec:rel_works}
\textbf{Image-Level OOD Detection}: In the image-level out-of-distribution (OOD) detection task, an image must be classified \textit{either} as in-distribution (ID) \textit{or} OOD. Early approaches to this task rely on distinguishing between ID and OOD based on the confidence score of the classifier \cite{Hendrycks2017, Liang2018, pmlr-v162-hendrycks22a, liu2020energy, pmlr-v162-wei22d, Hsu_2020_CVPR}. Alternative approaches measure distances in the feature space of the classifier \cite{lu2024learning, ming2023how, sun2022, NEURIPS2018_abdeb6f5} or leverage representation and activation statistics \cite{sastry2020, xiao2021reallyneedlearnrepresentations, Tack2020}. Other approaches rely on generative models \cite{Gao_2023_ICCV, Graham_2023_CVPR, liu2023, 10.1007/978-3-319-59050-9_12, zong2018deep, Li_2023_CVPR} or leverage likelihoods-based scores \cite{Jie2019, Xiao2020}. However, this task formulation does not account for real-world settings in which \textit{one} image may contain multiple objects which need separate ID or OOD assessment.
\\
\textbf{Object OOD Detection}: In the object OOD detection task, each object predicted by an object detector needs to be classified either as ID or OOD. Traditional approaches to this task \cite{DBLP:conf/cvpr/WuD23, 11093101, Wu2023_Iccv, Du2022, Xuefeng2022b} require training of the object detector together with additional modules to separate ID objects from OOD objects.
To avoid retraining the detector itself, \texttt{SAFE} \cite{Wilson2023} proposes a post-hoc solution that only trains an additional MLP to classify objects as ID or OOD.
However, this still requires training on the full ID dataset for each newly deployed object detector.\\
\textbf{Foundation Models for OOD Detection}: These limitations have motivated the use of vision-language foundation models, such as \texttt{CLIP} \cite{radford2021learningtransferablevisualmodels}, for few-shot and zero-shot OOD detection. While in the task of image-level OOD detection, this direction is extensively explored  \cite{Miyai2025, Wang_2023_ICCV, Esmaeilpourcharandabi2022, jiang2024negative, Ming2022, Zhou2022CoOp, chen2023plot, NEURIPS2023_f0606b88}, research on object OOD detection remains limited \cite{Li_2026_CVPR, zhang2025runaobjectleveloutofdistributiondetection, Nguyen2026}. Specifically, the few-shot method \texttt{RUNA} \cite{zhang2025runaobjectleveloutofdistributiondetection} uses two pretrained image encoders to extract global image features and regional object features, which are combined with text embeddings obtained from the predicted class labels to compute an OOD score for each object. The few-shot method \texttt{UNI-OOD} \cite{Li_2026_CVPR} improves over \texttt{RUNA} by using two identical pre-trained image and text encoder pairs to extract features of the target object and its background separately, which are then used to calculate an OOD score. However, both methods still require few-shot fine-tuning of custom modules on top of the pre-trained \texttt{CLIP} \cite{radford2021learningtransferablevisualmodels} image and text encoders.

The zero-shot object OOD detection method \texttt{RONIN} \cite{Nguyen2026} therefore proposes a retraining-free approach.
It leverages an off-the-shelf diffusion model to inpaint the object regions predicted by the object detector, conditioned on the corresponding predicted class labels.
After obtaining the inpainted objects, \texttt{RONIN} \cite{Nguyen2026} measures the text-to-image similarities using \texttt{CLIP} \cite{radford2021learningtransferablevisualmodels} between the text prompt and both the inpainted and original object, as well as the visual similarity between the original and inpainted object. 
These three similarities are used to calculate an OOD score. 
However, the use of diffusion-based inpainting leads to high computational cost, limiting its applicability for practical deployment.
Furthermore, OOD assessment in feature space can be sensitive to detector localization errors and variations in object appearance variability.

We address these limitations by proposing \ours, a post-hoc zero-shot object OOD detection method that requires no retraining for different object detectors or datasets.
Instead of relying on feature-space similarity, \ours performs OOD assessment at the more robust prediction level while requiring lower computational cost than an iterative diffusion-based inference.

%% file: sec/03_methods.tex
\section{Methods}
\label{sec:methods}
\begin{figure*}[t!]
    \centering
    \input{figures/method}
    \caption{Overview of \textbf{our proposed architecture for zero-shot object OOD detection.} Shown is an example, where the object detector predicts objects of class indices $s=1,4\text{ and }10$, which are the $R=3$ unique predicted classes.}
\label{fig:method}
\end{figure*}
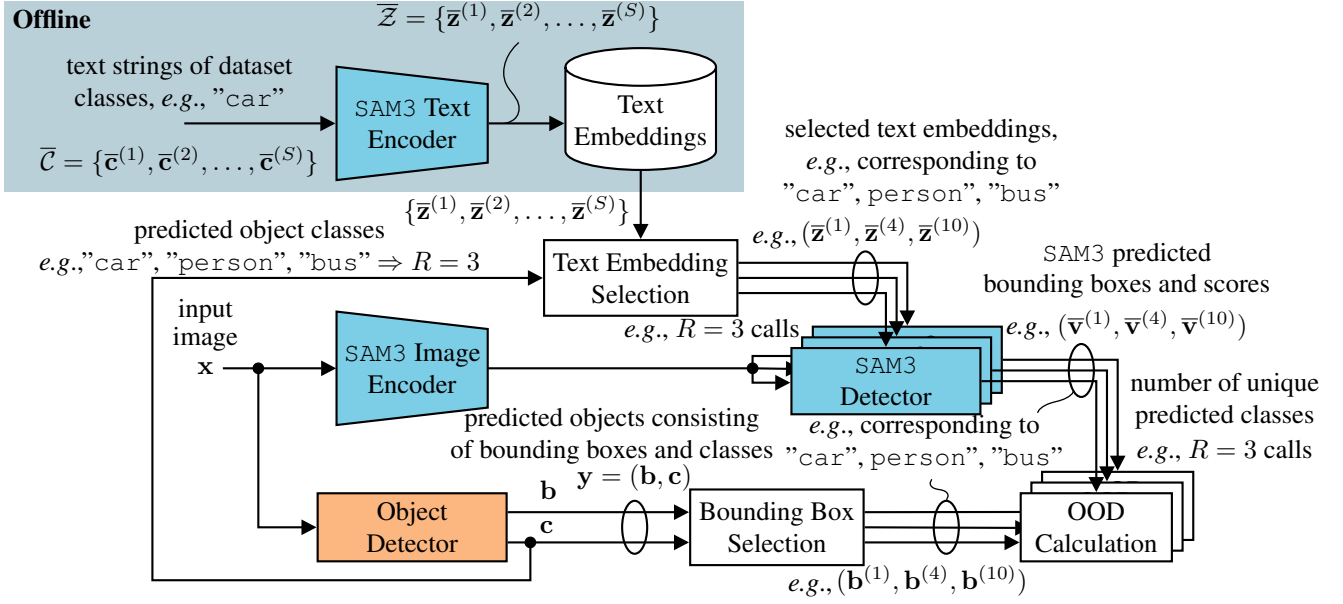

In this section, we describe our method \ours for zero-shot object out-of-distribution (OOD) detection.
The overall architecture of the proposed method is illustrated in \cref{fig:method}.
First, we describe how the object predictions produced by the object detector are used to prompt \texttt{SAM3-image} and obtain class-wise predictions.
We then detail how we use resulting \texttt{SAM3} detections to identify OOD objects among the object detector predictions.

\subsection{Prompting \textbf{\texttt{SAM3}} with Predicted Classes (Ours)}
\textbf{Object Detector Inference}: The input image $\mathbf{x} \in \mathbb{I}^{H \times W \times C}$, with $\mathbb{I}=[0,1]$ and $H$, $W$, and $C$ denoting the image height, width, and number of channels, respectively, is processed by both the object detector and the \texttt{SAM3} image encoder.
The object detector outputs predicted objects $\mathbf{y} = (\mathbf{b}, \mathbf{c})$, consisting of bounding boxes $\mathbf{b} = (\mathbf{b}_m)$ and object classes $\mathbf{c} = (c_m)$.
Here, $m \in \mathcal{M} = \{1,2,\dots,M\}$ denotes the object index, and $M$ is the number of predicted objects.
Each predicted object class $c_m=s$ is represented as a class text string $\overline{\mathbf{c}}^{(s)}$, \eg, \texttt{car}, and each bounding box is defined as $\mathbf{b}_m \in \mathcal{I}^2$ with pixel positions set $\mathcal{I} = \{(1,1),\dots,(H,W)\}$. Here, $s \in \mathcal{S}=\{1,\dots,S\}$ denotes the class index and $S$ the number of ID classes.\\
\textbf{Text Embedding Selection}: The predicted object classes $\mathbf{c}=(c_m)$ are then used to select the text embeddings of the corresponding class strings. These embeddings are precomputed \textit{offline} by the \texttt{\texttt{SAM3}} text encoder using the set $\overline{\mathcal{C}}=\{\overline{\mathbf{c}}^{(1)}, \dots, \overline{\mathbf{c}}^{(S)}\}$ of all in-distribution (ID) dataset class text strings. The resulting set of embeddings $\overline{\mathcal{Z}}= \{\overline{\mathbf{z}}^{(1)}, \overline{\mathbf{z}}^{(2)}, \dots, \overline{\mathbf{z}}^{(S)}\}$, $\overline{\mathbf{z}}^{(s)}\in \mathbb{R}^d$, is stored and used during inference. Note that $d$ is the dimensionality of the embedding space.
As an example, suppose that the object detector predicts $M=7$ objects of the $R=3$ classes \texttt{car}, \texttt{person}, and \texttt{bus}, corresponding to the class text strings $\overline{\mathbf{c}}^{(1)}$, $\overline{\mathbf{c}}^{(4)}$, and $\overline{\mathbf{c}}^{(10)}$, respectively, where $R$ denotes the number of unique predicted classes.
The text embedding selection step then retrieves the corresponding precomputed embeddings $\overline{\mathbf{z}}^{(1)}$, $\overline{\mathbf{z}}^{(4)}$, and $\overline{\mathbf{z}}^{(10)}$ to prompt the \texttt{SAM3} detector.\\
\textbf{\texttt{SAM3} Detector Inference}: The \texttt{SAM3} detector is inferred once for each unique predicted class $R$.
Thus, in the example above, the \texttt{SAM3} detector is inferred $R=3$ times.
For each prompted text embedding of class $s$, \texttt{SAM3} outputs $\overline{\mathbf{v}}^{(s)} = (\overline{\mathbf{b}}^{(s)}, \overline{\mathbf{u}}^{(s)})$, with bounding boxes $\overline{\mathbf{b}}^{(s)} = (\overline{\mathbf{b}}_{\overline{m}}^{(s)}) \in \mathcal{I}^{\overline{M}^{(s)}\times2}$ and confidence scores $\overline{\mathbf{u}}^{(s)} = (\overline{{u}}_{\overline{m}}^{(s)})\in [0,1]^{\overline{M}^{(s)}}$. 
Here, $\overline{m} \in \overline{\mathcal{M}}^{(s)} = \{1,\dots,\overline{M}^{(s)}\}$ indexes the \texttt{SAM3} predictions for class $s$ and $\overline{M}^{(s)}$ denotes their number.
We further define $\overline{\mathcal{Y}}_{\overline{m}}^{(s)}$ to be the set of all pixels inside bounding box $\overline{\mathbf{b}}_{\overline{m}}^{(s)}$ predicted by \texttt{SAM3}.

\subsection{OOD Detection (Ours)}
\textbf{Bounding Box Selection}: After obtaining class-wise predictions from \texttt{SAM3}, the corresponding bounding boxes of the object detector for OOD calculation need to be retrieved. This is done by the bounding box selection step (see \cref{fig:method}).
Using the object detector predictions $\mathbf{b}$ and $\mathbf{c}$, the bounding box selection step groups the predicted boxes $\mathbf{b}_m$ by their predicted class label ${c_m}$, yielding class-specific bounding boxes  $\mathbf{b}^{(s)} = (\mathbf{b}^{(s)}_{\hat{m}})$ corresponding to class $s$, \eg, $\mathbf{b}^{(1)}$, $\mathbf{b}^{(4)}$, and $\mathbf{b}^{(10)}$ for the above example of $R=3$ and $s \in \{1,4,10\}$. Here, $\hat{m} \in \mathcal{M}^{(s)} = \{1,\dots,M^{(s)}\}$ is the index for predicted bounding boxes corresponding to class $s$ and $M^{(s)}$ the number of predicted bounding boxes corresponding to class $s$. Here, we define $\mathcal{Y}_{\hat{m}}^{(s)}$ as the set of all pixels inside bounding box $\mathbf{b}^{(s)}_{\hat{m}}$.
The object detector bounding boxes and \texttt{SAM3} predictions corresponding to the same class, \eg, $\mathbf{b}^{(1)}$ and $\overline{\mathbf{v}}^{(1)}$ corresponding to $\overline{c}^{(1)}=\texttt{car}$, are then used to identify OOD objects predicted by the object detector, cf.\ \cref{fig:method_highlevel}.\\
\textbf{OOD Calculation}: First, the intersection over union (IoU)
\begin{equation}
    IoU(\mathbf{b}^{(s)}_{\hat{m}}, \overline{\mathbf{b}}^{(s)}_{\overline{m}}) = \frac{|\mathcal{Y}_{\hat{m}}^{(s)} \cap \overline{\mathcal{Y}}_{\overline{m}}^{(s)}|}{|\mathcal{Y}_{\hat{m}}^{(s)} \cup \overline{\mathcal{Y}}_{\overline{m}}^{(s)}|}
\end{equation}
is calculated between bounding box pixels of all objects predicted by the object detector, $\mathcal{Y}_{\hat{m}}^{(s)}$, and \texttt{SAM3}, $\overline{\mathcal{Y}}_{\overline{m}}^{(s)}$, that correspond to the same class $s$.
We then apply an IoU threshold $\theta^{\mathrm{IoU}} \in [0,1]$ and define
\begin{equation}
\label{equ:iou_threshold}
    \overline{\mathcal{M}}_{\hat{m}}^{(s)}
    =
    \{
        \overline{m} \in \overline{\mathcal{M}}^{(s)}
        \,|\,
        {IoU}\!(\mathbf{b}^{(s)}_{\hat{m}}, \overline{\mathbf{b}}^{(s)}_{\overline{m}})
        \geq
        \theta^{\mathrm{IoU}}
    \}.
\end{equation}
For each object detector prediction with index $\hat{m}$, the set $\overline{\mathcal{M}}_{\hat{m}}^{(s)}$ contains all \texttt{SAM3} predictions of the same class $s$ whose bounding boxes sufficiently overlap with the predicted object.
Then, an object detector prediction with index $\hat{m}$ is marked as ID (\textcolor{green}{\cmark}) if at least one of the \texttt{SAM3} prediction confidences $\overline{u}^{(s)}_{\overline{m}}$, $\overline{m} \in \overline{\mathcal{M}}^{(s)}_{\hat{m}}$ (\ref{equ:iou_threshold}), exceeds a score threshold $\theta^{\mathrm{score}} \in [0,1]$, with
\begin{equation}
\label{equ:score_thresh}
    \mathcal{M}_{\mathrm{ID}}^{(s)}
=
\{
    \hat{m} \in \mathcal{M}^{(s)}
    \,|\, \exists \ \overline{m} \in \overline{\mathcal{M}}_{\hat{m}}^{(s)}:
    \overline{u}_{\overline{m}}^{(s)} 
    \ge
    \theta^{\mathrm{score}}
\},
\end{equation}
being the set of all ID objects corresponding to class $s$.
Consequently, all predicted objects that are not marked as ID are considered OOD (\textcolor{tu3}{\xmark}).
The set of OOD predictions for class $s$ is therefore given by
\begin{equation}
    \mathcal{M}_{\mathrm{OOD}}^{(s)}
    =
    \mathcal{M}^{(s)} \setminus \mathcal{M}_{\mathrm{ID}}^{(s)}.
\end{equation}
Intuitively, our method flags objects as OOD when they are not detected by \texttt{SAM3} at the same location under the same class prompt.
In this way, we leverage the broad semantic knowledge of \texttt{SAM3} to verify whether the image content is aligned with the class predicted by the object detector.
For example, in \cref{fig:method_highlevel}, the object detector incorrectly classifies an \texttt{elephant} as \texttt{cow}, however, \texttt{SAM3}, when prompted with \texttt{cow}, does not predict an object at that location because the visual content does not correspond to a \texttt{cow}.
\textit{In contrast to previous \texttt{CLIP}-based approaches \cite{Nguyen2026, Li_2026_CVPR, zhang2025runaobjectleveloutofdistributiondetection} which measure the similarity between the predicted class and the image content of the predicted bounding box in an abstract feature space, our approach performs this comparison directly at the prediction level.} 
This has two advantages. 
First, OOD detection errors are easier to interpret, as illustrated later by qualitative results in \Cref{subsec:qulatitave_results}.
This facilitates refinement for future real-world applications, \eg, by enriching class prompts with descriptive adjectives, which is supported by \texttt{SAM3}.
Second, prediction-level agreement provides a more robust OOD signal than feature-space similarity, whose embeddings inherently entangle object appearance such as size and lighting and are not robust against detector bounding-box localization errors.

%% file: figures/method.tex
\pgfdeclarelayer{background}
\pgfdeclarelayer{bg2}
\pgfsetlayers{bg2,background,main}
\begin{tikzpicture}
	\coordinate (left) at (-2.5,0) {} {} {} {};
	
	\node (input) at (left) {$\mathbf{x}$};
	
	\node[align=center, trapezium, rotate=-90, draw, trapezium angle=77.5, minimum height=20mm, minimum width=15mm,
    inner sep=0.1pt,
    outer sep=0.1pt, thick, fill=tu4, right = 1.5cm of input, anchor = south, trapezium stretches=true] (image_encoder) {};
	\node at ($(image_encoder.west)!0.5!(image_encoder.east)$) [ align=center] {\texttt{SAM3} Image \\ Encoder};
	
	\draw[thick,-{Triangle[length=2mm,width=2mm]}] (input) -- (image_encoder);
	
	\node[align=center, trapezium, rotate=-90, draw, trapezium angle=77.5, minimum height=20mm, minimum width=15mm, inner sep=10pt, thick, fill=tu4, above = 2.6cm of image_encoder.west, anchor=center, yshift = -0cm, trapezium stretches=true] (text_encoder) {};
	\node at ($(text_encoder.west)!0.5!(text_encoder.east)$) [ align=center] {\texttt{SAM3} Text \\ Encoder};
	
	\node[align=center, cylinder, shape border rotate=90, aspect=0.25, anchor=north, draw, right = 1.0 of text_encoder.north, minimum width=1.0cm, minimum height=1.2cm, thick, fill=white] (text_embeddings) {Text \\ Embeddings};
	
	\node[align=center, rectangle, draw, below = 0.9 of text_embeddings.south, minimum width=2.cm, minimum height=1.0cm, thick, anchor=north] (embedding_selection) {Text Embedding \\ Selection};
	
	\node[left = 0.1cm of text_encoder.south,yshift=-0.5cm] (text_input) {$\overline{\mathcal{C}}=\{ \overline{\mathbf{c}}^{(1)}, \overline{\mathbf{c}}^{(2)}, \dots, \overline{\mathbf{c}}^{(S)}\}$};
    
    \node[left = -1.4cm of text_embeddings.west, yshift=1.4cm] (embeddings) {$\overline{\mathcal{Z}}= \{\overline{\mathbf{z}}^{(1)}, \overline{\mathbf{z}}^{(2)}, \dots, \overline{\mathbf{z}}^{(S)}\}$};

    \node[left = -1.0cm of text_embeddings.west, yshift=-1.15cm] (embeddings2) {$\{\overline{\mathbf{z}}^{(1)}, \overline{\mathbf{z}}^{(2)}, \dots, \overline{\mathbf{z}}^{(S)}\}$};

    \draw [line width=0.5, xshift=2cm] plot [smooth, tension=1] coordinates { (embeddings.south) ($(embeddings.south) + (-0.2, -0.4)$) ($(text_encoder.north) + (0.4, 0.2)$) ($(text_encoder.north) + (0.2cm, 0.0)$)};
	
	\draw[thick,-{Triangle[length=2mm,width=2mm]}] ($(text_encoder.south) + (-2cm,0cm)$) -- (text_encoder);
	
	\node[align=center, rectangle, draw, above = -0.17 of image_encoder.north, fill=tu4, minimum width=2.50cm, minimum height=0.8cm, thick, anchor=west, xshift = 4cm] (detector) {\texttt{SAM3} \\ Detector};

    \begin{pgfonlayer}{background}
    \node[align=center, rectangle, draw, above right = 0.4cm of detector.center, fill=tu4, minimum width=2.50cm, minimum height=0.8cm, thick, anchor=center] (detector2) {\texttt{SAM3} \\ Detector};
    
    \node[align=center, rectangle, draw, above right = 0.2cm of detector.center, fill=tu4, minimum width=2.50cm, minimum height=0.8cm, thick, anchor=center] (detector3) {\texttt{SAM3} \\ Detector};
    \end{pgfonlayer}

    \draw[thick,-{Triangle[length=2mm,width=2mm]}] ($(embedding_selection.east) + (0.0cm,-0.2cm)$) -| ($(detector.north) + (0.0cm,0.0cm)$);
    \draw[thick,-{Triangle[length=2mm,width=2mm]}] ($(embedding_selection.east) + (0.0cm,0.0cm)$) -| ($(detector3.north) + (0.0cm,0.0cm)$);
    \draw[thick,-{Triangle[length=2mm,width=2mm]}] ($(embedding_selection.east) + (0.0cm,0.2cm)$) -| ($(detector2.north) + (0.0cm,0.0cm)$);
	
	\draw[thick,-{Triangle[length=2mm,width=2mm]}] (text_encoder.north) -- (text_embeddings);
	
	\draw[thick,-{Triangle[length=2mm,width=2mm]}] (text_embeddings) -- (embedding_selection);

    \coordinate (split3) at ($(image_encoder.north) + (3.5cm, 0.0)$);
	\filldraw (split3) circle (2pt);  
    
    \begin{pgfonlayer}{bg2}
    \draw[thick] (image_encoder) --  (split3);
    
    \draw[thick,-{Triangle[length=2mm,width=2mm]}] (split3) |- ($(detector2.west) + (0.0cm, 0.05cm)$);
    \draw[thick,-{Triangle[length=2mm,width=2mm]}] (split3) |- ($(detector.west) + (0.0cm, -0.02cm)$);
    \draw[thick,-{Triangle[length=2mm,width=2mm]}] (split3) -- (detector3);
    \end{pgfonlayer}
	

	
	\node [above = 0.2cm of text_input.north, anchor = south, yshift=0.cm, align=center, thick] (detect_text)  {text strings of dataset \\ classes, \eg, "\texttt{car}"};
	
	\node [above = -0.1cm of input.north, anchor = south, yshift=0.cm, align=center, thick] (detect_text)  {input \\[-0.3ex] image};
	
	\node[align=center, rectangle, draw, below = 1 of image_encoder.east, fill=tu22, minimum width=2.50cm, minimum height=0.8cm, thick, anchor=north] (od) {Object \\ Detector};
	
	\coordinate (split1) at ($(input) + (0.7cm, 0.0)$);
	\filldraw (split1) circle (2pt);  
	
	\draw[thick,-{Triangle[length=2mm,width=2mm]}] (split1) |- (od);

    \node[align=center, rectangle, draw, right = 2.4 of od, minimum width=2.0cm, minimum height=1.0cm, thick, anchor=west] (bbox_selection) {Bounding Box \\ Selection};
	
	\node[align=center, rectangle, draw, right = 2.05 of bbox_selection, minimum width=2.0cm, minimum height=0.8cm, thick, anchor=west, fill=white] (ood) {OOD \\ Calculation};

    \begin{pgfonlayer}{background}
    \node[align=center, rectangle, draw, above right = 0.4 of ood.center, minimum width=2.0cm, minimum height=0.8cm, thick, anchor=center, fill=white] (ood3) {OOD \\ Calculation};
    
    \node[align=center, rectangle, draw, above right = 0.2 of ood.center, minimum width=2.0cm, minimum height=0.8cm, thick, anchor=center, fill=white] (ood2) {OOD \\ Calculation};
    \end{pgfonlayer}

    \draw[thick,-{Triangle[length=2mm,width=2mm]}] (detector) -| (ood);
    \draw[thick,-{Triangle[length=2mm,width=2mm]}] (detector3) -| (ood2);
    \draw[thick,-{Triangle[length=2mm,width=2mm]}] (detector2) -| (ood3);
	
	\draw[thick,-{Triangle[length=2mm,width=2mm]}] ($(od.east) + (0.0cm,0.2cm)$) -- ($(bbox_selection.west) + (0.0cm,0.2cm)$);

    \draw[thick,-{Triangle[length=2mm,width=2mm]}] ($(od.east) + (0.0cm,-0.2cm)$) -- ($(bbox_selection.west) + (0.0cm,-0.2cm)$);
    
    \draw[thick,-{Triangle[length=2mm,width=2mm]}] ($(bbox_selection.east) + (0.0cm,-0.2cm)$) -- ($(ood.west) + (0.0cm,-0.2cm)$);
    
    \begin{pgfonlayer}{background}
    \draw[thick,-{Triangle[length=2mm,width=2mm]}] ($(bbox_selection.east) + (0.0cm,-0.0cm)$) -- ($(ood2.west) + (0.0cm,-0.15cm)$);

    \draw[thick,-{Triangle[length=2mm,width=2mm]}] ($(bbox_selection.east) + (0.0cm,+0.2cm)$) -- ($(ood3.west) + (0.0cm,-0.08cm)$);
    \end{pgfonlayer}

	
	
	
	\coordinate (split2) at ($(od.east) + (0.3cm, -0.2)$);
	\filldraw (split2) circle (2pt);  
	
	\draw[thick,-{Triangle[length=2mm,width=2mm]}] (split2) |- ($(split2) + (-5.0, -0.5)$) |- ($(embedding_selection.west) + (0, 0.0)$);

    \node[right = 0.7cm of embedding_selection.east, yshift=0.6cm] (desc_emb) {$(\overline{\mathbf{z}}^{(1)}, \overline{\mathbf{z}}^{(4)}, \overline{\mathbf{z}}^{(10)})$};
    \node[left = -0.2cm of desc_emb.west, anchor=east, yshift=-0.1cm] (desc_emb_eg) {\eg,};

    \node[ellipse, draw, thick, right = 1.5cm of embedding_selection, minimum height = 0.7cm] (elipse_emb) {};

    \node [above = -0.1cm of desc_emb.north, anchor = south, xshift=0.4cm, align=center, thick] (selec_emb_text)  {selected text embeddings, \\ \eg, corresponding to \\ "\texttt{car}", \texttt{person}", "\texttt{bus}"};

    \node[right = 0.3cm of od.east, yshift=0.5cm] (desc_bbox) {${\mathbf{b}}$};
    \node[right = 0.3cm of od.east, yshift=0.0cm] (desc_label) {${\mathbf{c}}$};
    
    \node[ellipse, draw, thick, right = 1.5cm of od, minimum height = 0.7cm] (elipse_obj) {};
    
    \node[above = -0.01cm of elipse_obj, yshift=0.0cm] (obj_label) {${\mathbf{y}} = ({\mathbf{b}},{\mathbf{c}})$};

    \node [above = -0.25cm of obj_label.north, anchor = south, xshift=-0.3cm, align=center, thick] (selec_emb_text)  {predicted objects consisting \\ of bounding boxes and classes};

    \node [left = 0.7 cm of embedding_selection.west, anchor = east, yshift=0.4cm, align=center, thick] (example_classes)  {predicted object classes \\ \eg,"\texttt{car}", "\texttt{person}", "\texttt{bus}" $\Rightarrow R=3$};

    \node[right = -0.5cm of bbox_selection.east, yshift=-0.7cm] (des_boxes) {$({\mathbf{b}}^{(1)}, {\mathbf{b}}^{(4)}, {\mathbf{b}}^{(10)})$};
    \node[left = -0.2cm of des_boxes.west, anchor=east, yshift=-0.1cm] (des_boxes_eg) {\eg,};

    \node[ellipse, draw, thick, right = 0.9cm of bbox_selection, minimum height = 0.7cm] (elipse_box) {};

    \node [above = 1.01cm of des_boxes.north, anchor = south, xshift=-0.1cm, align=center, thick] (selec_emb_text)  {\eg, corresponding to \\ "\texttt{car}", \texttt{person}", "\texttt{bus}"};

    \node[ellipse, draw, thick, right = 1.0cm of detector3, minimum height = 0.7cm] (elipse_box_sam3) {};

    \node[above = -0.1cm of elipse_box_sam3.north, xshift=0.6cm] (des_boxes_sam3) {\eg, $(\overline{\mathbf{v}}^{(1)}, \overline{\mathbf{v}}^{(4)}, \overline{\mathbf{v}}^{(10)})$};

    \node [above = -0.1cm of des_boxes_sam3.north, anchor = south, xshift=-0.0cm, align=center, thick] (selec_emb_text_sam3)  {\texttt{SAM3} predicted \\ bounding boxes and scores};

    \draw [line width=0.5, xshift=2cm] plot [smooth, tension=1] coordinates { ($(selec_emb_text.north) + (1.5, -0.1)$) ($(selec_emb_text.north) + (1.6, 0.0)$) ($(elipse_box_sam3.south) + (-0.2, -0.0)$) ($(elipse_box_sam3.south) + (-0.1cm, 0.1)$)};

    \draw [line width=0.5, xshift=2cm] plot [smooth, tension=1] coordinates { ($(selec_emb_text.south) + (0.1, -0.0)$) ($(selec_emb_text.south) + (0.1, -0.1)$) ($(elipse_box.north) + (0.0, 0.1)$) ($(elipse_box.north) + (0.0cm, 0.0)$)};

    \begin{pgfonlayer}{background}
    \node[
        rectangle,
        thick,
        dashed,
        fill=tu62!60,
        fit=(text_input) (text_encoder) (text_embeddings) (embeddings),
        inner xsep=0.35cm,
        inner ysep=-0.02cm,
        yshift=-0.1cm,
        label={[anchor=north west]north west:\textbf{Offline}}
    ] (offline_box) {};
    \end{pgfonlayer}

    \node[left = -0.25cm of detector.west, yshift=0.67cm] (unqique_sam_calls) {\eg, $R=3$ calls};

    \node[right = -0.6cm of ood.east, yshift=1.0cm] (unqique_ood_calls) {\eg, $R=3$ calls};

    \node[above = -0.1cm of unqique_ood_calls.north, align=center, xshift=0cm] (unqique_ood_calls_des) {number of unique \\ predicted classes};




\end{tikzpicture}

%% file: sec/04_experimental_setup.tex
\section{Experimental Setup}
\label{sec:exp_setup}
\textbf{Data}: Following the standard evaluation setup of object OOD detection \cite{Du2022, Xuefeng2022b, Nguyen2026}, we use Pascal-VOC \cite{everingham2010pascal} and BDD100K \cite{Yu2018b} as in-distribution (ID) datasets and evaluate out-of-distribution (OOD) detection performance on two subsets from MS-COCO \cite{lin2014microsoft} and OpenImages \cite{Kuznetsova2020}.
These subsets are constructed such that their test images contain no label overlap with the corresponding ID datasets. Additionally, as employed by our main baseline \cite{Nguyen2026}, we use the first $400$ images of Pascal-VOC, OpenImages, and MS-COCO and the first $200$ images of BDD100K to balance object count between ID and OOD datasets.\\
\textbf{Metrics}: OOD detection performance is evaluated using the area under the receiver operating characteristic curve (AuROC) and the false positive rate at a true positive rate of $95\%$ (FPR95). Following the standard convention \cite{Du2022, Xuefeng2022b, Nguyen2026}, a true positive denotes an ID object correctly classified as ID, whereas a false positive denotes an OOD object incorrectly classified as ID. We obtain the ROC curves for AuROC calculation as typical in the field. For our particular method, we obtain the ROC curves by varying the confidence score threshold $\theta^{\mathrm{score}}$ in (\ref{equ:score_thresh}), while we firmly use a particular IoU threshold $\theta^{\mathrm{IoU}}$ (\ref{equ:iou_threshold}) and keep it fixed. We do not report average precision (AP) of the object detector, since our method does not modify the object detector. Accordingly, AP results are only provided in Appendix A.\\
\textbf{Models}: Following \cite{Nguyen2026, Xuefeng2022b}, for all reported methods, we use \texttt{DeformableDETR} \cite{Zhu2021} as object detector pre-trained on the respective ID datasets. For \ours, we also use the off-the-shelf \texttt{SAM3-image} model \cite{carion2026sam3segmentconcepts}. We used the \texttt{facebook/sam3} weights and standard preprocessing\footnote{available at \href{https://huggingface.co/facebook/sam3}{https://huggingface.co/facebook/sam3}}.\\
\textbf{Baselines}: Our main baseline is the so-far SOTA zero-shot object OOD method \texttt{RONIN} \cite{Nguyen2026}. Due to a lack of zero-shot methods for object OOD detection, we follow \cite{Nguyen2026} and additionally compare our method with image-level zero-shot OOD detection approaches applied to the object task, by cropping each object and treating it as an image. Specifically, \texttt{CLIP}-based \texttt{ODIN} \cite{Liang2018} and \texttt{EnergyScore} \cite{liu2020energy}, \texttt{Mahalanobis} \cite{xiao2021reallyneedlearnrepresentations}, \texttt{KNN} \cite{sun2022}, \texttt{MCM} \cite{Ming2022}, \texttt{CLIPN} \cite{Wang_2023_ICCV}, \texttt{TAG} \cite{Liu2024}, and \texttt{GL-MCM} \cite{Miyai2025} are reported. We also report two basic training-based methods \texttt{VOS} \cite{Du2022} and \texttt{SIREN} \cite{Xuefeng2022b}.

%% file: sec/05_experimental_results.tex
\section{Experimental Results and Discussion}
\label{sec:results}
\subsection{Main Results}
\begin{table*}[t!]
    \centering
    \caption{\textbf{Object OOD detection results} on ID datasets Pascal-VOC and BDD100K and on OOD datasets MS-COCO and OpenImages. We report AuROC (\%) and FPR95 (\%). We adopt baseline results from \cite{Nguyen2026}. Best results in \textbf{bold}, second-best results underlined.}
    \input{tables/main_results}
    \label{tab:main_results}
\end{table*}
In \cref{tab:main_results}, we compare our zero-shot method for object OOD detection to so-far state-of-the-art (SOTA) zero-shot methods across two ID datasets, namely Pascal-VOC (top segment) and BDD100K (bottom segment). We measure OOD detection performance using AuROC (\%, higher is better) and FPR95 (\%, lower is better). An overall average rank over both metrics and both OOD datasets is reported (cf.\ \cite{Saijo2025}). The baseline results are adopted from \cite{Nguyen2026}.

We observe that \texttt{EnergyScore} \cite{liu2020energy} achieves a decent 3rd rank on the Pascal-VOC ID dataset (top), while \texttt{Mahalanobis} secures a rank $3.1$ on the BDD100K ID dataset (bottom). Both, however, fall behind when applied to the respective other ID dataset. For both ID datasets, Pascal-VOC (top) and BDD100K (bottom), the overall strongest zero-shot object OOD baseline is \texttt{RONIN} \cite{Nguyen2026}, with an average rank of $2.0$ on Pascal-VOC and $2.6$ on BDD100K. \textit{Our \ours zero-shot object OOD detection method, however, excels consistently across both ID datasets, both OOD datasets, and both metrics, achieving new SOTA performance with an overall average rank of $1.0$.} Results using the full test sets can be found in Appendix B.

In \cref{tab:iou_ablation}, we conduct a sensitivity analysis of the IoU threshold $\theta^{\mathrm{IoU}}$, which is used in the OOD calculation to pre-select \texttt{SAM3} predictions (\ref{equ:iou_threshold}).
We report results using Pascal-VOC as the ID dataset and MS-COCO and OpenImages as OOD datasets, and measure OOD detection performance using AuROC and FPR95.
We observe that our method is largely insensitive to the choice of the IoU threshold $\theta^{\mathrm{IoU}}$.
However, on our datasets we observe a tendency towards better performance for a low $\theta^{\mathrm{IoU}}$.
Interestingly, \textit{all} threshold configurations outperform the strongest baseline \texttt{RONIN} \cite{Nguyen2026}, which achieved an average rank of $2.0$ with Pascal-VOC as ID dataset, as shown in \cref{tab:main_results}.
\begin{table}[t!]
    \centering
    \caption{\textbf{Sensitivity analysis of our \ours} w.r.t.\ the IoU threshold $\theta^{\mathrm{IoU}}$ (\ref{equ:iou_threshold}) on Pascal-VOC (ID) and MS-COCO and OpenImages (both OOD). We report AuROC (\%) and FPR95 (\%).}
    \input{tables/big_ablation}
    \label{tab:iou_ablation}
\end{table}
Although $\theta^{\mathrm{IoU}}=0.1$ yields the best performance in \cref{tab:iou_ablation}, selecting the smallest threshold may not generalize well to other test configurations.
The reason is as follows: In the established evaluation setup for object OOD detection \cite{Du2022, Xuefeng2022b, Wilson2023, Li_2026_CVPR, Nguyen2026, zhang2025runaobjectleveloutofdistributiondetection}, ID and OOD objects do not co-occur within the same image.
Under this assumption, weakly localized \texttt{SAM3} predictions can still help identify ID objects and reduce false negatives, without the risk of simultaneously classifying an OOD object in the same image as ID.
In more general open-world settings, however, where ID and OOD objects may co-occur, a very low IoU threshold would increase the risk of false positives by accepting insufficiently localized matches. \textit{Accordingly, for a broader generalization of our method, we propose the suboptimal $\theta^{\mathrm{IoU}}=0.2$ for thresholding the \texttt{SAM3} predictions in (\ref{equ:iou_threshold}). Note that with the better choice of $\theta^{\mathrm{IoU}}=0.1$, our results in \cref{tab:main_results} would have been even stronger.} More performance details on the influence of $\theta^{\mathrm{IoU}}$ are in Appendix C and results of a more realistic open-world setting in Appendix D.

\begin{figure}[t!]
    \centering
    \input{figures/performance_speed}
    \caption{\textbf{Performance-speed comparison} of our \ours (\textcolor{tu3}{red}) to the so-far SOTA baseline \texttt{RONIN} \cite{Nguyen2026} (\textcolor{tu5}{blue}) on an \texttt{NVIDIA A100 40GB} with ID datset Pascal-VOC and OOD datasets MS-COCO (\protect\tikz[baseline=-0.6ex]{\protect\draw[black, fill=black] (0,0) circle (2pt);}) and OpenImages (\protect\tikz[baseline=-0.6ex]{\protect\draw[black, fill=white, thick] (0,0) circle (2pt);}). Our method employs $\theta^{\mathrm{IoU}}=0.2$ (\ref{equ:iou_threshold}). The \textcolor{tu91}{green} symbols mark our approach limited to $R\le3$.}
\label{fig:performance_speed}
\end{figure}

In \cref{fig:performance_speed}, we compare object OOD detection performance and runtime per processed frame of our proposed \ours with the strongest zero-shot baseline \texttt{RONIN} \cite{Nguyen2026}, which achieved strong ranks $2.0$ and $2.6$ with Pascal-VOC and BDD100K ID data, respectively, as reported in \cref{tab:main_results}. The three marks for \texttt{RONIN} \cite{Nguyen2026}, from left to right, correspond to configurations with $5$, $10$, and $20$ diffusion steps, respectively, where the strongest latter is the configuration reported in \cref{tab:main_results}. The runtime is measured using an \texttt{NVIDIA A100 40GB} and the OOD detection performance is measured by AuROC (top) and FPR95 (bottom). Runtime is measured only for the OOD detection method, excluding the object detector, since both methods employ the identical \texttt{DeformableDETR} \cite{Zhu2021} detector. As the runtime per frame can vary for both methods, we report the average runtime over all frames with at least on predicted object in the respective ID and OOD datasets. We observe that \ours is processed substantially faster than even the fastest \texttt{RONIN} \cite{Nguyen2026} configuration with five diffusion steps.
Specifically, our method achieves a runtime of $0.1$\,s per frame, compared to $0.48$\,s for the fastest \texttt{RONIN} \cite{Nguyen2026} configuration.
At the same time, it achieves significantly better object OOD detection performance than even the strongest \texttt{RONIN} \cite{Nguyen2026} configuration.

In \cref{fig:dataset_statistics}, we analyze the amount of unique predicted classes of the \texttt{DeformableDETR} \cite{Zhu2021} object detector in the OOD datasets MS-COCO (top) and OpenImages (bottom).
The statistics are shown in green for the object detector trained on the Pascal-VOC ID dataset and in purple for the object detector trained on the BDD100K ID dataset.
We observe that for both OOD datasets and both ID-trained models, the amount $R$ of unique predicted classes is relatively low, with most images containing either $R=0$ or $R=1$ predicted classes.
This comes with positive consequences for our proposed \ours approach. It indicates that the computational overhead of inferring the \texttt{SAM3} detector $R$ times per image is limited in practice, since the \texttt{SAM3} detector is not at all inferred in 37...76\% of the frames ($R=0$) and is inferred a single time only in 19...53\% of the frames ($R=1$).
For real-world deployment of our \ours,
the computational cost could be bounded by enforcing a maximum number of unique prompted classes, \eg, $R \leq 3$. If the object detector predicts more than $3$ unique classes, we prompt only the $R=3$ classes with the lowest average detector confidence across their corresponding predictions.
All remaining predictions are considered ID.
The performance of our approach limited to a maximum of $R \leq 3$ \texttt{SAM3} calls per frame is shown in green color \protect\tikz[baseline=-0.6ex]{\protect\draw[tu91, fill=tu91] (0,0) circle (2pt);}, \protect\tikz[baseline=-0.6ex]{\protect\draw[tu91, fill=white, thick] (0,0) circle (2pt);} in \cref{fig:performance_speed}, always nearby the respective red markers \protect\tikz[baseline=-0.6ex]{\protect\draw[tu3, fill=tu3] (0,0) circle (2pt);}, \protect\tikz[baseline=-0.6ex]{\protect\draw[tu3, fill=white, thick] (0,0) circle (2pt);} of our unbounded method (any $R$). This limitation comes with some performance loss, however, our computationally limited \ours method still excels \texttt{RONIN} in all cases.

In \cref{tab:runtime_components} we further analyze the runtime (ms) on an \texttt{NVIDIA A100 40GB} of the \texttt{SAM3-image} \cite{carion2026sam3segmentconcepts} used in our zero-shot \ours. Specifically, we report the isolated runtime of the individual components used during inference, the \texttt{SAM3} image encoder and the \texttt{SAM3} detector, when inferred with $R=1$, $R=2$, and $R=3$ prompts.
We observe that roughly $60 \%$ of the runtime reported in \cref{fig:performance_speed} is caused by the \texttt{SAM3} image encoder. This runtime cannot be reduced by bounding $R$, since the image encoder always has to be inferred \textit{once} per image. 
Additionally, we observe that the runtime of the \texttt{SAM3} detector does not scale linearly with the number of prompts $R$, since the \texttt{SAM3} detector can be inferred in a batched manner.
Furthermore, as shown in \cref{fig:dataset_statistics}, only a few samples are affected when bounding $R$. 
Together, these factors explain why bounding the method to $R\le 3$ yields only a very minor speedup in \cref{fig:performance_speed}. Adding $63.4$ ms for inferring the \texttt{SAM3} image encoder and the worst case $53.1$ ms for $R=3$ batched calls of the \texttt{SAM3} detector, yields $116.5$ ms, which is the \textit{frame-individual} worst case of the $R\leq 3$ constrained green method's runtime in \cref{fig:performance_speed}. In Appendix E, we provide additional comparisons on computational complexity.\\
In Appendix F, we additionally provide comparisons with state-of-the-art \textit{few-shot} object OOD detection methods.
\begin{figure}[t!]
    \centering
    \input{figures/dataset_statistics}
    \caption{\textbf{Analysis of amount $R$ of unique predicted classes} per OOD dataset, using the \texttt{DeformableDETR} \cite{Zhu2021} object detector.}
\label{fig:dataset_statistics}
\end{figure}
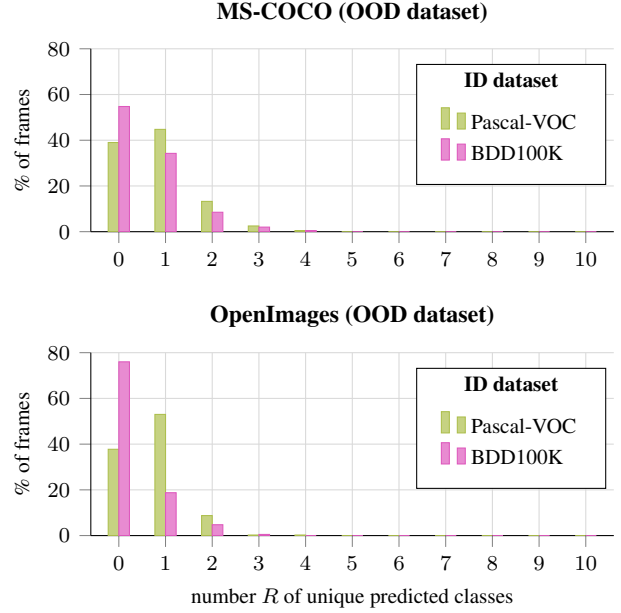
\begin{table}[t!]
    \centering
    \caption{\textbf{Runtime of the single components} of the \texttt{SAM3-image} for our \ours in ms on an \texttt{Nvidia A100 40GB}.}
    \setlength{\tabcolsep}{4pt}
    \input{tables/runtime_components}
    \label{tab:runtime_components}
\end{table}
\subsection{Sensitivity Analysis Ablation}
Now, we investigate the sensitivity of the so-far SOTA \texttt{RONIN} \cite{Nguyen2026} and our \ours method to object detector bounding-box localization errors and object appearance.

In \cref{tab:loc_error}, we analyze the robustness of our method, \ours, and the strongest baseline, \texttt{RONIN} \cite{Nguyen2026}, to bounding-box localization errors of the object detector with Pascal-VOC as ID and MS-COCO as OOD dataset.
To induce localization errors, we follow the approach of B\"ar et al.\ \cite{Baer2023a} and perturb the bounding box coordinates predicted by the object detector according to some noise strength $\epsilon \geq 0$. Specifically, we add uniformly sampled noise to the bounding box coordinates (upper left, lower right), with the noise scaled by the corresponding bounding box width and height. More details on generating the erroneous bounding boxes and the exact definition of the noise strength can be found in Appendix G.
After obtaining the erroneous object detector bounding boxes, both object OOD detection methods are evaluated as usual. The performance of \texttt{RONIN} \cite{Nguyen2026} with no noise ($\epsilon=0$) differs slightly from the values reported in \cref{tab:main_results}, as we re-evaluated their strongest method for the robustness analysis using their published source code.  

We observe that \texttt{RONIN} \cite{Nguyen2026} degrades continuously with simulated growing object detector bounding-box localization error, leading to an increase in FPR of $6.8 \%$ absolute for $\epsilon=0.5$. Our \ours, however, shows very robust performance on both metrics across all evaluated error strengths $\epsilon$. \textit{This supports our claim that performing OOD detection directly at prediction level, rather than in the feature space of models like \texttt{CLIP} \cite{radford2021learningtransferablevisualmodels}, is advantageous due to its higher robustness to bounding-box localization errors.}

\begin{table}[t!]
    \centering
    \caption{\textbf{Robustness to bounding box localization errors}. Comparison of our \ours and \texttt{RONIN} \cite{Nguyen2026} baseline under object detector localization errors on Pascal-VOC as ID dataset and MS-COCO as OOD dataset. We report AuROC (\%) / FPR95 (\%).}
    \input{tables/loc_error}
    \label{tab:loc_error}
\end{table}

In \cref{tab:object_appearance}, we analyze how different object-appearance factors, such as lighting and object size, affect our method, \ours, and the strongest baseline, \texttt{RONIN} \cite{Nguyen2026}.
For lighting, we compute the median luminance $Y \in [0,1]$ within each bounding box predicted by \texttt{DeformableDETR} \cite{Zhu2021}.
We then assign objects to high-lighting ($Y \geq 0.3$) and low-lighting ($Y < 0.3$) categories.
For object size, we follow the MS-COCO definition \cite{lin2014microsoft} and categorize objects as small, medium, or large.
Finally, we compute the object OOD detection metrics separately for the ID and OOD objects in each category.

We observe that both methods perform worse under low-lighting conditions than under high-lighting conditions.
However, the absolute performance drop is somewhat larger for \texttt{RONIN} \cite{Nguyen2026} than for our \ours.
This effect is even more pronounced for object size. While \ours still performs well on small objects, achieving $90.76\%$ AuROC, \texttt{RONIN} \cite{Nguyen2026} degrades to near-chance performance with $51.09\%$ AuROC. 
\textit{This further supports our claim that our method provides a more robust OOD signal with respect to object appearance than feature-space similarity employed by \texttt{RONIN} \cite{Nguyen2026} and other baselines.}

\begin{table}[t!]
    \centering
    \caption{\textbf{Robustness to different object appearances.}  Comparison of our \ours and \texttt{RONIN} \cite{Nguyen2026} baseline under different object appearances on Pascal-VOC as ID dataset and MS-COCO as OOD dataset. We report AuROC (\%) / FPR95 (\%).} 
    \input{tables/object_appearance}
    \label{tab:object_appearance}
\end{table}

\subsection{Qualitative Results and Failure Analysis}
\label{subsec:qulatitave_results}
In Appendix H we provide qualitative results showing several success and failure cases and a failure analysis of our proposed \ours.

%% file: tables/main_results.tex
\begin{tabular}{
    c|l|c|
    S[table-format=2.2, table-number-alignment=center]
    S[table-format=2.2, table-number-alignment=center]
    S[table-format=2.2, table-number-alignment=center]
    S[table-format=2.2, table-number-alignment=center]|
    S[table-format=2.1, table-number-alignment=center]
}
    \toprule[.9pt]
    & & & \multicolumn{4}{c|}{Out-of-distribution (OOD) datasets} & \multicolumn{1}{c}{} \\
    In-distribution & Method & Zero-shot?
    & \multicolumn{2}{c}{MS-COCO}
    & \multicolumn{2}{c|}{OpenImages}
    & \multicolumn{1}{c}{Average} \\
    (ID) dataset & & &
    \multicolumn{1}{c}{AuROC $\uparrow$} &
    \multicolumn{1}{c}{FPR95 $\downarrow$} &
    \multicolumn{1}{c}{AuROC $\uparrow$} &
    \multicolumn{1}{c|}{FPR95 $\downarrow$} &
    \multicolumn{1}{c}{rank $\downarrow$} \\
    \midrule
    \multirow{13}{*}{Pascal-VOC}
    & \texttt{VOS} \cite{Du2022}           &\xmark & 88.75 & 48.15 & 83.65 & 54.63 & 7.3 \\
    & \texttt{SIREN} \cite{Xuefeng2022b}         &\xmark & 78.68 & 64.70 & 75.12 & 66.69 & 11.8 \\
    & \texttt{ODIN} \cite{Liang2018}          &\cmark & 88.22 & 41.65 & 86.46 & 55.87 & 6.8 \\
    & \texttt{Energy Score} \cite{liu2020energy}  &\cmark & 90.26 & 29.48 & 91.24 & 24.57 & 3.0 \\
    & \texttt{Mahalanobis} \cite{xiao2021reallyneedlearnrepresentations}   &\cmark & 83.26 & 63.30 & 87.11 & 45.22 & 8.2 \\
    & \texttt{KNN} \cite{sun2022}           &\cmark & 85.52 & 58.56 & 84.62 & 45.00 & 7.2 \\
    & \texttt{MCM} \cite{Ming2022}           &\cmark & 83.15 & 62.47 & 81.45 & 71.52 & 11.2 \\
    & \texttt{CLIPN} \cite{Wang_2023_ICCV}         &\cmark & 85.45 & 43.09 & 89.31 & 41.74 & 5.2 \\
    & \texttt{TAG} \cite{Liu2024}           &\cmark & 78.35 & 61.03 & 87.69 & 48.48 & 8.8 \\
    & \texttt{OLE} \cite{ding2024zero}           &\cmark & 86.13 & 54.23 & 88.07 & 49.13 & 6.5 \\
    & \texttt{GL-MCM} \cite{Miyai2025}        &\cmark & 78.64 & 78.56 & 82.89 & 69.35 & 12.0 \\
    & \texttt{RONIN} \cite{Nguyen2026}         &\cmark & \underline{92.31} & \underline{25.36} & \underline{93.10} & \underline{18.91} & 2.0 \\
    & \ours (ours)               &\cmark & \bfseries 97.52 & \bfseries 6.56 & \bfseries 94.62 & \bfseries 16.52 & \bfseries 1.0 \\
    \midrule
    \multirow{13}{*}{BDD100K}
    & \texttt{VOS} \cite{Du2022}           &\xmark & 78.34 & 65.45 & 80.42 & 59.23 & 8.8 \\
    & \texttt{SIREN} \cite{Xuefeng2022b}         &\xmark & 89.37 & 42.86 & 91.78 & 37.97 & 5.5 \\
    & \texttt{ODIN} \cite{Liang2018}          &\cmark & 55.18 & 96.51 & 57.87 & 95.56 & 12.3 \\
    & \texttt{Energy Score} \cite{liu2020energy}  &\cmark & 74.49 & 71.75 & 73.09 & 53.33 & 9.5 \\
    & \texttt{Mahalanobis} \cite{xiao2021reallyneedlearnrepresentations}   &\cmark & 90.74 & 31.75 & \underline{93.81} & \underline{23.33} & 3.1 \\
    & \texttt{KNN} \cite{sun2022}           &\cmark & 86.58 & 35.87 & 92.75 & 31.11 & 4.8 \\
    & \texttt{MCM} \cite{Ming2022}           &\cmark & 55.82 & 95.56 & 57.05 & {92.22} & 11.5 \\
    & \texttt{CLIPN} \cite{Wang_2023_ICCV}         &\cmark & 92.14 & 28.49 & 85.78 & 44.76 & 4.8 \\
    & \texttt{TAG} \cite{Liu2024}           &\cmark & 90.50 & 46.30 & 76.67 & 55.30 & 7.5 \\
    & \texttt{OLE} \cite{ding2024zero}           &\cmark & 78.14 & 57.14 & 85.36 & 40.00 & 7.5 \\
    & \texttt{GL-MCM} \cite{Miyai2025}        &\cmark & 50.82 & 96.83 & 58.88 & 94.44 & 12.3 \\
    & \texttt{RONIN} \cite{Nguyen2026}         &\cmark & \underline{92.90} & \underline{26.03} & 91.90 & \underline{23.33} & 2.6 \\
    & \ours (ours)               &\cmark & \bfseries 98.46 & \bfseries 4.55 & \bfseries 98.40 & \bfseries 5.00 & \bfseries 1.0 \\
    \bottomrule[.9pt]
\end{tabular}

%% file: tables/big_ablation.tex
\begin{tabular}{l|cc|cc}
    \toprule[.9pt]
      $\theta^{\mathrm{IoU}}$ & \multicolumn{2}{c|}{MS-COCO} & \multicolumn{2}{c}{OpenImages} \\
     & AuROC $\uparrow$ & FPR95 $\downarrow$ & AuROC $\uparrow$ & FPR95 $\downarrow$ \\
    \midrule
        0.10  & 97.60 & 6.11 & 94.88 & 15.62 \\
        0.15 & 97.54 & 6.56 & 94.72 & 15.62 \\
        0.20  & 97.52 & 6.56 & 94.62 & 16.52 \\
        0.25 & 97.51 & 6.56 & 94.51 & 17.72 \\
        0.30  & 97.40 & 7.47 & 94.40 & 17.72 \\
    \bottomrule[.9pt]
\end{tabular}

%% file: figures/performance_speed.tex
\begin{tikzpicture}

\pgfplotsset{
    roninplot/.style={
        width=7.0cm,
        height=4.8cm,
        xmin=0.05, xmax=1.1,
        xtick={0.1,0.3,0.5,0.7,0.9,1.1},
        grid=major,
        grid style={gray!35},
        tick label style={font=\footnotesize},
        label style={font=\footnotesize},
        title style={font=\bfseries\small},
        xlabel style={yshift=0.4ex},
        ylabel style={yshift=-0.6ex}
    }
}

\begin{axis}[
    roninplot,
    name=auc_coco,
    ylabel={AuROC (\%)},
    ymin=89, ymax=99,
    ytick={88,90,92,94,96,98},
    xlabel={runtime (s)}
]
\addplot[
    color=tu5,
    thick,
    mark=*,
    mark size=2.4pt,
    mark options={fill=tu5, draw=tu5}
] coordinates {
    (0.462,90.35)
    (0.649,91.10)
    (1.018,92.31)
};

\addplot[
    color=tu5,
    thick,
    mark=*,
    mark size=2.4pt,
    mark options={fill=white, draw=tu5}
] coordinates {
    (0.462,92.32)
    (0.649,92.80)
    (1.018,93.10)
};

\addplot[
    only marks,
    color=tu3,
    thick,
    mark=*,
    mark size=2.4pt,
    mark options={fill=tu3, draw=tu3}
] coordinates {
    (0.105,97.76)
};

\addplot[
    only marks,
    color=tu3,
    thick,
    mark=*,
    mark size=2.4pt,
    mark options={fill=white, draw=tu3}
] coordinates {
    (0.105,95.30)
    };

\addplot[
    only marks,
    color=tu3,
    thick,
    mark=*,
    mark size=2.4pt,
    mark options={fill=tu91, draw=tu91}
] coordinates {
    (0.101,96.64)
};

\addplot[
    only marks,
    color=tu81,
    thick,
    mark=*,
    mark size=2.4pt,
    mark options={fill=white, draw=tu91}
] coordinates {
    (0.101,93.75)
    };

\node[font=\footnotesize, text=tu5, anchor=west] at (axis cs:0.48,93.7) {\textbf{\texttt{RONIN} \cite{Nguyen2026}}};

\node[font=\footnotesize, text=tu3, anchor=west] at (axis cs:0.15,97.05) {\textbf{\ours (ours)}};
\node[font=\footnotesize, text=tu91, anchor=west] at (axis cs:0.15,96.05) {\textbf{\ours, $R \le 3$ (ours)}};

\end{axis}

\begin{axis}[
    roninplot,
    name=fpr_coco,
    at={(auc_coco.south)},
    anchor=north,
    yshift=-1.7cm,
    ylabel={FPR95 (\%)},
    ymin=0, ymax=32,
    ytick={0,5,10,15,20,25,30},
    xlabel={runtime (s)}
]
\addplot[
    color=tu5,
    thick,
    mark=*,
    mark size=2.4pt,
    mark options={fill=tu5, draw=tu5}
] coordinates {
    (0.462,29.94)
    (0.649,29.07)
    (1.018,25.36)
};

\addplot[
    color=tu5,
    thick,
    mark=*,
    mark size=2.4pt,
    mark options={fill=white, draw=tu5}
] coordinates {
    (0.462,20.00)
    (0.649,20.20)
    (1.018,18.91)
};

\addplot[
    only marks,
    color=tu3,
    thick,
    mark=*,
    mark size=2.4pt,
    mark options={fill=tu3, draw=tu3}
] coordinates {
    (0.105,5.88)
    };

\addplot[
    only marks,
    color=tu3,
    thick,
    mark=*,
    mark size=2.4pt,
    mark options={fill=white, draw=tu3}
] coordinates {
    (0.105,15.02)
};

\addplot[
    only marks,
    color=tu3,
    thick,
    mark=*,
    mark size=2.4pt,
    mark options={fill=tu91, draw=tu91}
] coordinates {
    (0.101,7.47)
};

\addplot[
    only marks,
    color=tu3,
    thick,
    mark=*,
    mark size=2.4pt,
    mark options={fill=white, draw=tu91}
] coordinates {
    (0.101,17.42)
    };

\node[font=\footnotesize, text=tu5, anchor=west] at (axis cs:0.48,24.3) {\textbf{\texttt{RONIN} \cite{Nguyen2026}}};

\node[font=\footnotesize, text=tu3, anchor=west] at (axis cs:0.15,10.6) {\textbf{\ours (ours)}};
\node[font=\footnotesize, text=tu91, anchor=west] at (axis cs:0.15,6.6) {\textbf{\ours, $R \le 3$ (ours)}};

\end{axis}
\end{tikzpicture}

%% file: figures/dataset_statistics.tex
\begin{tikzpicture}

\pgfplotsset{
    histplot/.style={
        width=8.5cm,
        height=4.0cm,
        ybar,
        bar width=4pt,
        xmin=0, xmax=10,
        ymin=0, ymax=80,
        xtick={0,1,...,10},
        ytick={0,20,40,60,80},
        ylabel={\% of frames},
        grid=major,
        grid style={gray!30},
        tick label style={font=\footnotesize},
        label style={font=\footnotesize},
        title style={font=\bfseries\small},
        axis x line*=bottom,
        axis y line*=left,
        tick align=outside,
        enlarge x limits=0.06,
        legend style={
            draw=none,
            fill=none,
            font=\footnotesize
        },
        legend cell align={left}
    }
}

\begin{axis}[
    histplot,
    name=mscoco,
    title={MS-COCO (OOD dataset)},
    legend entries={Pascal-VOC,BDD100K},
    legend to name=combinedlegend
]

\addplot[
    draw=tu91,
    fill=tu91!70,
    bar shift=-2pt
] coordinates {
    (0,39.00)
    (1,44.75)
    (2,13.25)
    (3,2.50)
    (4,0.50)
    (5,0.00)
    (6,0.00)
    (7,0.00)
    (8,0.00)
    (9,0.00)
    (10,0.00)
};

\addplot[
    draw=tu111,
    fill=tu111!70,
    bar shift=2pt
] coordinates {
    (0,54.75)
    (1,34.25)
    (2,8.50)
    (3,2.00)
    (4,0.50)
    (5,0.00)
    (6,0.00)
    (7,0.00)
    (8,0.00)
    (9,0.00)
    (10,0.00)
};

\end{axis}

\begin{axis}[
    histplot,
    name=openimages,
    at={(mscoco.south)},
    anchor=north,
    yshift=-1.6cm,
    title={OpenImages (OOD dataset)},
    xlabel={number $R$ of unique predicted classes},
]

\addplot[
    draw=tu91,
    fill=tu91!70,
    bar shift=-2pt
] coordinates {
    (0,37.75)
    (1,53.00)
    (2,8.75)
    (3,0.25)
    (4,0.25)
    (5,0.00)
    (6,0.00)
    (7,0.00)
    (8,0.00)
    (9,0.00)
    (10,0.00)
};

\addplot[
    draw=tu111,
    fill=tu111!70,
    bar shift=2pt
] coordinates {
    (0,76.00)
    (1,18.75)
    (2,4.75)
    (3,0.50)
    (4,0.00)
    (5,0.00)
    (6,0.00)
    (7,0.00)
    (8,0.00)
    (9,0.00)
    (10,0.00)
};

\end{axis}

\node[
    draw=black,
    fill=white,
    inner xsep=0.1pt,
    inner ysep=0.1pt,
    minimum width=1.0cm,
    minimum height=0.1cm,
    anchor=north east
] at ([xshift=-0.1cm,yshift=-0.2cm]mscoco.north east) {%
    \begin{tabular}{c}
        \footnotesize\textbf{ID dataset} \\[2pt]
        \pgfplotslegendfromname{combinedlegend}
    \end{tabular}
};

\node[
    draw=black,
    fill=white,
    inner xsep=0.1pt,
    inner ysep=0.1pt,
    minimum width=1.0cm,
    minimum height=0.1cm,
    anchor=north east
] at ([xshift=-0.1cm,yshift=-0.2cm]openimages.north east) {%
    \begin{tabular}{c}
        \footnotesize\textbf{ID dataset} \\[2pt]
        \pgfplotslegendfromname{combinedlegend}
    \end{tabular}
};

\end{tikzpicture}

%% file: tables/runtime_components.tex
\begin{tabular}{l|c|ccc}
    \toprule[.9pt]
    Component &\texttt{SAM3} &\multicolumn{3}{c}{\texttt{SAM3} detector} \\
    &image encoder &$R=1$ &$R=2$ &$R=3$ \\
    \midrule
    Runtime (ms) &$63.4$ &$36.0$ &$48.8$ &$53.1$   \\
    \bottomrule[.9pt]
\end{tabular}

%% file: tables/loc_error.tex
\begin{tabular}{l|cc}
    \toprule[.9pt]
    Localization &\ours &\texttt{RONIN} \\
    error $\epsilon$ &(ours) &\cite{Nguyen2026} \\
    \midrule
    $0$ &$97.52/6.56$ &$90.70/27.84$   \\
    $0.1$ &$97.51/6.58$ &$90.59/29.69$  \\
    $0.3$ &$97.54/6.58$ &$90.10/29.48$ \\
    $0.5$ &$97.54/6.58$ &$88.43/34.64$ \\
    \bottomrule[.9pt]
\end{tabular}

%% file: tables/object_appearance.tex
\begin{tabular}{l|cc}
    \toprule[.9pt]
    Object &\ours &\texttt{RONIN} \\
    appearance &(ours) &\cite{Nguyen2026} \\
    \midrule
     Lighting & &   \\
    \quad high ($Y \ge 0.3$) &$98.18/06.49$ &$90.20/30.88$  \\
     \quad low ($Y < 0.3$) &$94.85/10.78$ &$85.31/36.27$  \\
          \midrule
    Object size & &   \\
    \quad large &$98.29/03.72$ &$94.71/21.14$  \\
     \quad medium &$94.74/14.49$ &$74.65/52.17$  \\
     \quad small &$90.76/21.74$ &$51.09/86.96$  \\    
    \bottomrule[.9pt]
\end{tabular}

%% file: sec/06_conclusions.tex
\section{Conclusions}
\label{sec:conclusions}
In this paper, we introduce \ours, a zero-shot object out-of-distribution (OOD) detection method that efficiently leverages the \texttt{SAM3-image} foundation model.
\textit{Our method outperforms all baselines across both in-distribution (ID) datasets, both OOD datasets, and both metrics, while significantly requiring lower computational cost than the current SOTA method. Accordingly, our \ours method constitutes a new state-of-the-art in zero-shot object OOD detection.}
We further analyze and demonstrate the robustness of prediction-level OOD assessment compared to feature-space similarity.

%% file: app/sec/ap_results.tex
\section{AP Results of the Object Detector}

In \cref{tab:ap_results} we report the average precision (AP) metric (\%) of the \texttt{DeformableDETR} \cite{Zhu2021} object detector on the ID datasets Pascal-VOC \cite{everingham2010pascal} and BDD100K \cite{Yu2018b}. This object detector is used for evaluating all object out-of-distribution (OOD) detection methods.

%% file: app/sec/full_test_set.tex
\section{Results on Full Test Sets}
In \cref{tab:full}, we evalaute our \ours for object OOD detection using \textit{entire} Pascal-VOC and BDD100K as ID datasets and \textit{entire} MS-COCO and OpenImages as OOD datasets. We report AuROC and FPR95 metrics and observe that also on the entire test sets \ours outperforms \texttt{RONIN} \cite{Nguyen2026} \textit{by an even larger margin than on the official reported numbers} by \cite{Nguyen2026} in Tab.\ 1.

%% file: app/sec/abl_iou.tex
\section{Performance Influence of the IoU Threshold}
In \cref{tab:iou_threshold}, we analyze the effect of the IoU threshold $\theta^{\mathrm{IoU}}$ in (2) on object out-of-distribution (OOD) detection performance.
The results are obtained using in-distribution (ID) datasets Pascal-VOC \cite{everingham2010pascal} (top segment) and BDD100K \cite{Yu2018b} (bottom segment) and the OOD datasets MS-COCO \cite{lin2014microsoft} and OpenImages \cite{Kuznetsova2020}.
We measure object OOD detection performance using AuROC and FPR95.

Consistent with the results in Tab.\ 2 (main paper), object OOD performance increases with lower $\theta^{\mathrm{IoU}}$, achieving best performance across all metrics and datasets even at $\theta^{\mathrm{IoU}}=0$.
However, as discussed in Section 5.1, choosing a very small IoU threshold, such as $\theta^{\mathrm{IoU}}=0$, would not generalize well to other test configurations.
In particular, if ID and OOD objects co-occur in the same image, accepting weakly (or not at all) localized matches would substantially increase the FPR.
This is why we propose $\theta^{\mathrm{IoU}}=0.2$ for our main results in Tab.\ 1 (main paper). 
For fairness, however, future work solely evaluating on this particular setting, where ID and OOD objects cannot co-occur in the same image, should also report the performance of \ours with $\theta^{\mathrm{IoU}}=0$.

%% file: app/sec/openworld_setting.tex
\section{Evaluation in an Open-World Setup}
In \cref{tab:openworld}, we evaluate our proposed \ours on a more realistic open-world setup, where ID and OOD objects can occur within the same image. For this setup, we use the \texttt{DeformableDETR} \cite{Zhu2021} trained on Pascal-VOC as ID dataset and evaluate on the full MS-COCO \cite{lin2014microsoft} validation set. For the open-world , we introduce a new labeling protocol for ID and OOD objects.
\begin{table}[t!]
    \centering
    \caption{\textbf{Object detection performance} of the \texttt{DeformableDETR} \cite{Zhu2021} object detector on the in-distribution (ID) datasets Pascal-VOC \cite{everingham2010pascal} and BDD100K \cite{Yu2018b}. We report mean average precision (AP) (\%).}
    \input{app/tables/ap_results}
    \label{tab:ap_results}
\end{table}
\begin{table}[t!]
    \centering
    \caption{\textbf{Object OOD detection results} using \ours on Pascal-VOC and BDD100K as ID datasets and MS-COCO and OpenImages as OOD datasets. We report AuROC (\%) / FPR95 (\%).}
    \input{app/tables/full_test_set}
    \label{tab:full}
\end{table}
\begin{table}[t!]
    \centering
    \caption{\textbf{Performance influence of $\theta^{\mathrm{IoU}}$} in our \ours on Pascal-VOC and BDD100K as ID datasets and MS-COCO and OpenImages as OOD datasets. We report AuROC (\%) and FPR95 (\%). Best results in \textbf{bold}.}
    \input{app/tables/iou_threshold}
    \label{tab:iou_threshold}
\end{table}
First, we match \texttt{DeformableDETR} predictions to ground-truth objects based on their IoU.
A prediction is considered matched if its IoU with a ground-truth object is greater than $0.3$.
If a prediction is matched to a ground-truth object of the same class, we label it as ID. If it is matched to a ground-truth object of a class not being part of the ID dataset, we label it as OOD. In this evaluation for object OOD detection, we ignore all other predictions.
Under this protocol, ID and OOD objects can co-occur within the same image.
We then apply the object OOD methods as usual to the predicted objects to identify the OOD objects.
We observe that, as expected, the open-world setup is more challenging for both the baseline \texttt{RONIN} \cite{Nguyen2026} and our \ours, with both methods performing worse than in the Pascal-VOC MS-COCO setup in Tab.\ 1. \textit{However, our proposed \ours outperforms \texttt{RONIN} by an even larger margin in the more realistic setup, demonstrating the better practical applicability of our method.} Furthermore, our method with $\theta^{\mathrm{IoU}}=0.2$ outperforms the variant with $\theta^{\mathrm{IoU}}=0.0$, supporting our choice of this threshold for \textit{better generalization to realistic open-world setups.}
\begin{table}[t!]
    \centering
    \caption{\textbf{Object OOD detection results} on an open-world setup on the full MS-COCO validation set.  We report AuROC (\%) and FPR95 (\%). Best results in \textbf{bold}.}
    \input{app/tables/openworld}
    \label{tab:openworld}
\end{table}

%% file: app/tables/ap_results.tex
\begin{tabular}{l|cc}
    \toprule[.9pt]
    Dataset &Pascal-VOC &BDD100K \\
    \midrule
    AP (\%) &$60.6$ &$31.3$   \\
    \bottomrule[.9pt]
\end{tabular}

%% file: app/tables/full_test_set.tex
\begin{tabular}{l|c|c}
    \toprule[.9pt]
      Method & {MS-COCO} & {OpenImages} \\
    \midrule
        \multicolumn{3}{c}{ID dataset: Pascal-VOC} \\
        \midrule
        \texttt{RONIN} \cite{Nguyen2026}  & 89.67 / 30.30 & 91.43 / 25.19 \\
        \ours & \textbf{97.72} / \enspace\textbf{6.24} & \textbf{93.87} / \textbf{15.97} \\
        \midrule
        \multicolumn{3}{c}{ID dataset: BDD100K} \\
        \midrule
        \texttt{RONIN} \cite{Nguyen2026}  & 78.41 / 57.91 & 83.86 / 44.37  \\
        \ours & \textbf{97.99} / \enspace\textbf{5.25} & \textbf{98.47} / \enspace\textbf{3.26} \\
    \bottomrule[.9pt]
\end{tabular}

%% file: app/tables/iou_threshold.tex
\begin{tabular}{l|cc|cc}
    \toprule[.9pt]
      $\theta^{\mathrm{IoU}}$ & \multicolumn{2}{c|}{MS-COCO} & \multicolumn{2}{c}{OpenImages} \\
     & AuROC $\uparrow$ & FPR95 $\downarrow$ & AuROC $\uparrow$ & FPR95 $\downarrow$ \\
    \midrule
        \multicolumn{5}{c}{ID dataset: Pascal-VOC} \\
        \midrule
        0 & \textbf{98.14} & \textbf{5.20} & \textbf{96.18} & \textbf{12.01} \\
        0.1  & 97.60 & 6.11 & 94.88 & 15.62 \\
        0.2  & 97.52 & 6.56 & 94.62 & 16.52 \\
        0.3 & 97.40 & 7.47 & 94.40 & 17.72 \\
        \midrule
        \multicolumn{5}{c}{ID dataset: BDD100K} \\
        \midrule
        0 & \textbf{99.42} & \textbf{1.99} & \textbf{99.57} & \textbf{0.71} \\
        0.1  & 98.65 & 3.98 & 98.49 & 5.71 \\
        0.2  & 98.46 & 4.55 & 98.40 & 5.00 \\
        0.3  & 98.01 & 4.83 & 97.91 & 5.00 \\
    \bottomrule[.9pt]
\end{tabular}

%% file: app/tables/openworld.tex
\begin{tabular}{l|cc}
    \toprule[.9pt]
    Method &AuROC $\uparrow$ &FPR95 $\downarrow$ \\
    \midrule
    \texttt{RONIN} \cite{Nguyen2026} &$81.72$ &$51.21$   \\
    \ours ($\theta^{\mathrm{IoU}}=0.0$) &$92.15$ &$20.63$   \\
    \ours ($\theta^{\mathrm{IoU}}=0.2$) &$\mathbf{94.11}$ &$\mathbf{16.19}$   \\
    \bottomrule[.9pt]
\end{tabular}

%% file: app/sec/computational_complexity.tex
\section{Comparison of Computational Complexity}
In \cref{tab:flops}, we compare compuational complexity in terms of TFLOPs of the strongest zero-shot baseline \texttt{RONIN} \cite{Nguyen2026} and our \ours for the case of $R=1$, $R=2$, and $R=3$ unique prompted classes. For both methods, we use the configurations reported in Tab 1.
Specifically, \texttt{RONIN} uses $20$ diffusion steps and a diffusion-model input resolution of $512 \times 512$, whereas \ours uses an input resolution of $1008 \times 1008$. We observe that, for all values of $R$, \ours requires substantially fewer FLOPs.
Furthermore, the computational complexity of \texttt{RONIN} scales linearly with the number of prompted classes, since its main source of FLOPs is the diffusion model, which must be evaluated once for each additional class.
For \ours, in contrast, only the \texttt{SAM3} detector is evaluated additionally for each prompted class, which limits the increase in FLOPs.
\begin{table}[t!]
    \centering
    \caption{\textbf{Computational complexity} in terms of TFLOPs of \texttt{RONIN} \cite{Nguyen2026} and our \ours. Best results in \textbf{bold}. }
    \input{app/tables/flops}
    \label{tab:flops}
\end{table}

%% file: app/tables/flops.tex
\begin{tabular}{l|cccc}
    \toprule[.9pt]
    Method  &$R=1$ &$R=2$ &$R=3$ \\
    \midrule
    \texttt{RONIN} \cite{Nguyen2026} &$32.19$ &$64.39$ &$96.58$   \\
    \ours &\enspace$\mathbf{4.30}$ &\enspace$\mathbf{4.59}$ &\enspace$\mathbf{4.87}$   \\
    \bottomrule[.9pt]
\end{tabular}

%% file: app/sec/few_shot_methods.tex
\section{Comparison with Few-Shot Methods}
\begin{table*}[t!]
    \centering
    \caption{\textbf{Object OOD detection results} comparison of \ours with zero- and few-shot methods on Pascal-VOC and BDD100K as ID datasets and MS-COCO and OpenImages as OOD datasets. Baseline results are adopted from the respective papers. We report AuROC (\%) and FPR95 (\%) metrics. Best results in \textbf{bold}, second-best underlined.}
    \input{app/tables/few_shot}
    \label{tab:fewshot_threshold}
\end{table*}

In \cref{tab:fewshot_threshold}, we compare our \textit{zero-shot} \ours to state-of-the-art \textit{few-shot} methods for object OOD detection using Pascal-VOC and BDD100K as ID datasets and MS-COCO and OpenImages as OOD datasets.
We report AuROC and FPR95 metrics.
As expected, due to their advantageous access to task-specific data, the few-shot baseline \texttt{UNI-OOD} \cite{Li_2026_CVPR} outperforms the zero-shot baseline \texttt{RONIN} \cite{Nguyen2026} across all datasets and metrics. 
Our proposed \textit{zero-shot} method, \ours with $\theta^{\mathrm{IoU}}=0.2$, however, is competitive with the \textit{few-shot} \texttt{UNI-OOD} \cite{Li_2026_CVPR} and outperforms it in all settings with MS-COCO as the OOD dataset. 
Moreover, \ours with $\theta^{\mathrm{IoU}}=0.0$ surpasses \texttt{UNI-OOD} \cite{Li_2026_CVPR} even more generally, being best in $7$ out of $8$ dataset-metric combinations.

%% file: app/tables/few_shot.tex
\begin{tabular}{c|l|cccc}
    \toprule[.9pt]
      & & \multicolumn{4}{c}{Out-of-distribution (OOD) datasets} \\
    In-distribution & Method
    & \multicolumn{2}{c}{MS-COCO}
    & \multicolumn{2}{c}{OpenImages} \\
    (ID) dataset & & 
    \multicolumn{1}{c}{AuROC $\uparrow$} &
    \multicolumn{1}{c}{FPR95 $\downarrow$} &
    \multicolumn{1}{c}{AuROC $\uparrow$} &
    \multicolumn{1}{c}{FPR95 $\downarrow$} \\
    \midrule
    &\multicolumn{5}{c}{Few-shot methods} \\
    \cmidrule(lr){2-6}
    \multirow{5}{*}{Pascal-VOC}
        &\texttt{RUNA} \cite{zhang2025runaobjectleveloutofdistributiondetection} & 92.48 & 30.67 & 93.63 & 26.07 \\
        &\texttt{UNI-OOD} \cite{Li_2026_CVPR} & {95.07} & {22.24}  &\textbf{96.25} & \underline{14.30} \\
        \cmidrule(lr){2-6}
        &\multicolumn{5}{c}{Zero-shot methods} \\
    \cmidrule(lr){2-6}
        &\texttt{RONIN} \cite{Nguyen2026} & 92.31 & 25.36  & 93.10 & 18.91 \\
        &\ours ($\theta^{\mathrm{IoU}}=0.2$) (ours) & \underline{97.52} & \enspace\underline{6.56}  & {94.62} & {16.52} \\
        &\ours ($\theta^{\mathrm{IoU}}=0.0$) (ours) &\textbf{98.14} & \enspace\textbf{5.20} & \underline{96.18} & \textbf{12.01} \\
        \midrule
    &\multicolumn{5}{c}{Few-shot methods} \\
    \cmidrule(lr){2-6}
    \multirow{5}{*}{BDD100K}
        &\texttt{RUNA} \cite{zhang2025runaobjectleveloutofdistributiondetection} & 93.92 & 16.85 & 96.76 & \enspace9.95 \\
        &\texttt{UNI-OOD} \cite{Li_2026_CVPR} & {95.91} & {11.32}  & \underline{98.52} & \enspace\underline{3.68} \\
        \cmidrule(lr){2-6}
        &\multicolumn{5}{c}{Zero-shot methods} \\
    \cmidrule(lr){2-6}
        &\texttt{RONIN} \cite{Nguyen2026} & 92.90 & 26.03  & 91.90 & 23.33 \\
        &\ours ($\theta^{\mathrm{IoU}}=0.2$) (ours) & \underline{98.46} & \enspace\underline{4.55}  & {98.40} & \enspace{5.00} \\
         &\ours ($\theta^{\mathrm{IoU}}=0.0$) (ours) & \textbf{99.42} & \enspace\textbf{1.99} & \textbf{99.57} & \enspace\textbf{0.71} \\
    \bottomrule[.9pt]
\end{tabular}

%% file: app/sec/localization_error.tex
\section{Simulation of Bounding-Box Localization Error}
In the following, we describe how we use the approach of B\"ar et al.\ \cite{Baer2023a} to simulate bounding-box localization errors of the object detector.\\
\textbf{Preliminaries}: A bounding box predicted by the object detector is defined as $\mathbf{b}_{m} \in \mathcal{I}^2$ with pixel positions set $\mathcal{I} = \{(1,1),\dots,(H,W)\}$ and $H$ and $W$ being the image height and width, respectively. The bounding-box coordinates
\begin{equation}
    \mathbf{b}_{m} = ((h^{\mathrm{L}}_m,w^{\mathrm{L}}_m), (h^{\mathrm{R}}_m,w^{\mathrm{R}}_m)),
\end{equation}
are given by a top-left corner $(h^{\mathrm{L}}_m,w^{\mathrm{L}}_m)\in \mathcal{I}$ and a bottom-right corner $(h^{\mathrm{R}}_m,w^{\mathrm{R}}_m) \in \mathcal{I}$. Then, the size of the bounding box is given by $H_m \times W_m$ with
\begin{align}
    H_m = |h^{\mathrm{L}}_m- h^{\mathrm{R}}_m|,\\
    W_m = |w^{\mathrm{L}}_m- w^{\mathrm{R}}_m|,
\end{align}
where $H_m$ and $W_m$ denote the object height and width, respectively.\\
\textbf{Error Simulation}: To induce bounding-box localization errors, we perturb the object detector bounding boxes $\mathbf{b}_m$ using some noise coordinate factor $\epsilon \ge 0$, also referred to as noise strength. Depending on this noise strength, we obtain object-specific uniform distributions
\begin{align}
    U_h(-u_h, u_h), \text{ with } u_h = \frac{\epsilon}{2} H_m, \\
    U_w(-u_w, u_w), \text{ with } u_w = \frac{\epsilon}{2} W_m, 
\end{align}
which are used to perturb the bounding-box coordinates. To this end, we sample offsets
\begin{align}
    \Delta h^{\mathrm{L}}_m, \Delta h^{\mathrm{R}}_m  \sim U_h(-u_h, u_h), \\
    \Delta w^{\mathrm{L}}_m, \Delta w^{\mathrm{R}}_m  \sim U_w(-u_w, u_w),
\end{align}
from the object-specific uniform distribution for the height coordinates $\Delta h^{\mathrm{L}}_m, \Delta h^{\mathrm{R}}_m$ and width coordinates $\Delta w^{\mathrm{L}}_m, \Delta w^{\mathrm{R}}_m$. Finally, we obtain the erroneous bounding box
\begin{equation}
\begin{aligned}
    \tilde{\mathbf{b}}_m
    ={}& ((h^{\mathrm{L}}_m,w^{\mathrm{L}}_m)
    + (\Delta h^{\mathrm{L}}_m, \Delta w^{\mathrm{L}}_m), \\
    &\quad (h^{\mathrm{R}}_m,w^{\mathrm{R}}_m)
    + (\Delta h^{\mathrm{R}}_m, \Delta w^{\mathrm{R}}_m)),
\end{aligned}
\end{equation}
by adding the offsets to the original bounding-box coordinates predicted by the object detector. The offsets are limited in a way that no disturbed bounding box is extended towards pixel positions outside the image.

%% file: app/sec/qualitative_results.tex
\section{Qualitative Results and Failure Analysis}
\begin{figure*}[t!]
    \centering
    \input{app/figures/qualitative_results}
    \caption{\textbf{Qualitative results of our \ours method} for object OOD detection. In the top and middle row, success cases for the ID dataset Pascal-VOC (left) and the OOD datasets MS-COCO (middle) and OpenImages (right) are shown. In the bottom row, failure cases for the respective datasets are illustrated. Objects and labels are predicted by the \texttt{DeformableDETR} \cite{Zhu2021} and classified by our \ours as ID (\textcolor{green}{\cmark}) or OOD (\textcolor{tu3}{\xmark}).}
\label{fig:qualitative_results}
\end{figure*}

\textbf{Qualatative Results}: In \cref{fig:qualitative_results}, we show qualitative results of our \ours method. Illustrated are success cases (top and middle rows) and failure cases (bottom row) for Pascal-VOC ID objects (left), MS-COCO OOD objects (middle), and OpenImages OOD objects (right).
The shown labels in \cref{fig:qualitative_results} are the output of the object detector \texttt{DeformableDETR} and are then used to prompt \texttt{SAM3-image} \cite{carion2026sam3segmentconcepts} according to our proposed \ours method. For ID success cases (top and middle left), \texttt{SAM3-image} predicts bounding boxes at the corresponding object detector locations, correctly marking the objects as ID (\textcolor{green}{\cmark}).
Similarly, for OOD success cases (top and middle right), \texttt{SAM3} does \textit{not} predict corresponding bounding boxes, correctly marking the objects as OOD (\textcolor{tu3}{\xmark}).
In the ID failure case (bottom left), the ID object \texttt{chair} is marked as OOD (\textcolor{tu3}{\xmark}) because \texttt{SAM3-image} fails to predict an object at that location under the \texttt{chair} prompt. This is the case because the object is closer to an \texttt{armchair}.
In the OOD failure case on MS-COCO, the object detector predicts two objects, both labeled as \texttt{potted plant}. The right object is not confirmed by \texttt{SAM3} and is therefore correctly classified as OOD (\textcolor{tu3}{\xmark}). 
The left object in reality is a \texttt{pineapple}, not a \texttt{potted plant}. 
However, \texttt{SAM3} also predicts an object at that location under the \texttt{potted plant} prompt, and therefore \textit{incorrectly} classifies it as ID (\textcolor{green}{\cmark}).
In the OOD failure case on OpenImages (bottom right), the predicted object \texttt{potted plant} is not confirmed by a \texttt{SAM3} prediction and is therefore correctly classified as OOD (\textcolor{tu3}{\xmark}).
The second object predicted by the object detector is a \texttt{bottle} in the top-left region of the image. 
Since this \texttt{bottle} is also detected by \texttt{SAM3}, it is \textit{incorrectly} marked as ID (\textcolor{green}{\cmark}) under the evaluation protocol, where OOD images are assumed not to contain any ID objects.
Accordingly, in the standard evaluation protocol used in the field, all detected objects in an OOD image are expected to be marked as OOD.
However, a \texttt{bottle} is clearly visible in this image, suggesting a dataset-labeling issue rather than a true method failure, since our task is \textit{object} OOD detection.
We observed this type of error more frequently on OpenImages, which may partially explain the performance gap in Tab.\ 1 (main paper) compared to the other datasets.\\
\textbf{Failure Analysis}: For failure analysis, we use the operating point corresponding to the FPR95 metric with Pascal-VOC as the ID dataset and MS-COCO as the OOD dataset from Tab.\ 1.
At this operating point, the false negative rate (FNR) is $5.00\%$ and the false positive rate (FPR) is $6.56\%$.
For false negatives, \ie, ID objects that are marked as OOD because they are not detected by \texttt{SAM3}, we observe two types of failures.
The first type corresponds to errors made by the object detector, such as predicting the wrong class or predicting objects at locations where no object is present.
These incorrect predictions are not verified by \texttt{SAM3} and are therefore marked as OOD, although they originate from an ID image.
The second type corresponds to true method failures, where \texttt{SAM3} fails to confirm a real ID object.
The overall FNR of $5.00\%$ decomposes into $0.27\%$ false negatives where a real object is not confirmed by \texttt{SAM3} and $4.73\%$ false negatives caused by object-detector errors.
Accordingly, only a small fraction of false negatives are due to failures of our proposed \ours method.
From a broader error-detection perspective, where OOD objects are only one possible cause of detection errors, this further expands the possibilities of practical deployment of \ours.
Distinguishing between OOD objects and standard detector failures is typically simpler task for practical systems, compared to detecting erroneous detections. For false positives, the two types of errors occurring are the ones discussed in the qualitative results.

%% file: app/figures/qualitative_results.tex
\begin{tikzpicture}
\node[inner sep=0, minimum width=1.524cm, minimum height=0.752cm] (voc1) {\includegraphics[width=0.28\textwidth]{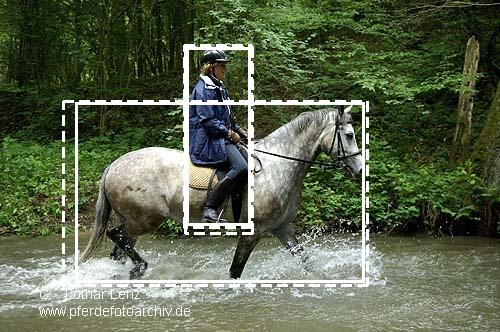}};

\node[inner sep=0, minimum width=1.524cm, minimum height=0.752cm, below = 0.72cm of voc1, anchor = north] (voc2) {\includegraphics[width=0.28\textwidth]{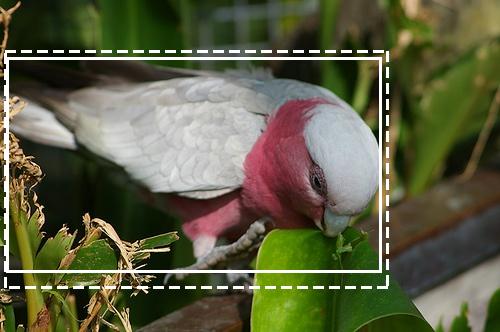}};

\node[inner sep=0, minimum width=1.524cm, minimum height=0.752cm, below = 0.3cm of voc2, anchor = north] (voc3) {\includegraphics[width=0.28\textwidth]{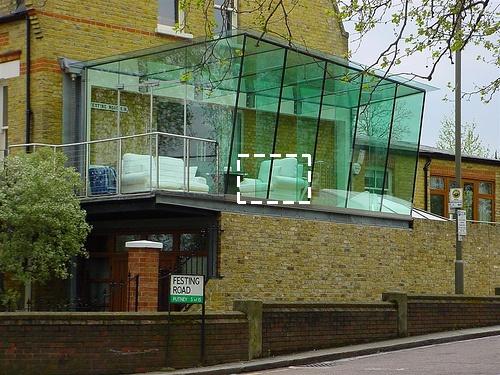}};


\node[inner sep=0, minimum width=1.524cm, minimum height=0.752cm, right = 0.7cm of voc1, anchor = west] (coco1) {\includegraphics[width=0.28\textwidth]{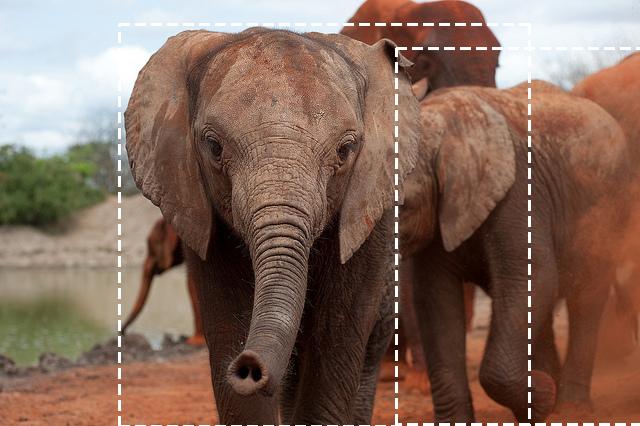}};

\node[inner sep=0, minimum width=1.524cm, minimum height=0.752cm, below = 0.3cm of coco1, anchor = north] (coco2) {\includegraphics[width=0.28\textwidth]{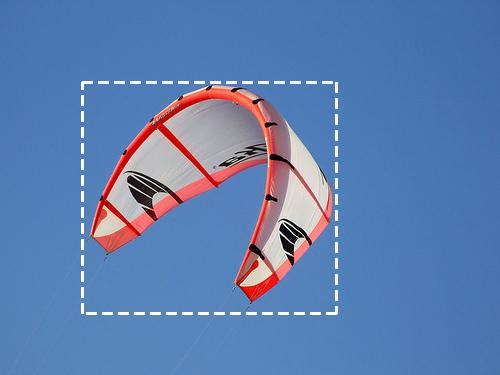}};

\node[inner sep=0, minimum width=1.524cm, minimum height=0.752cm, below = 0.3cm of coco2, anchor = north] (coco3) {\includegraphics[width=0.28\textwidth]{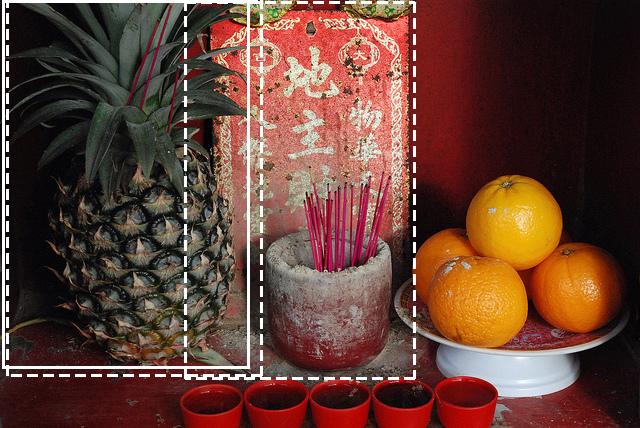}};


\node[inner sep=0, minimum width=1.524cm, minimum height=0.752cm, right = 0.7cm of coco1, anchor = west] (oi1) {\includegraphics[width=0.28\textwidth]{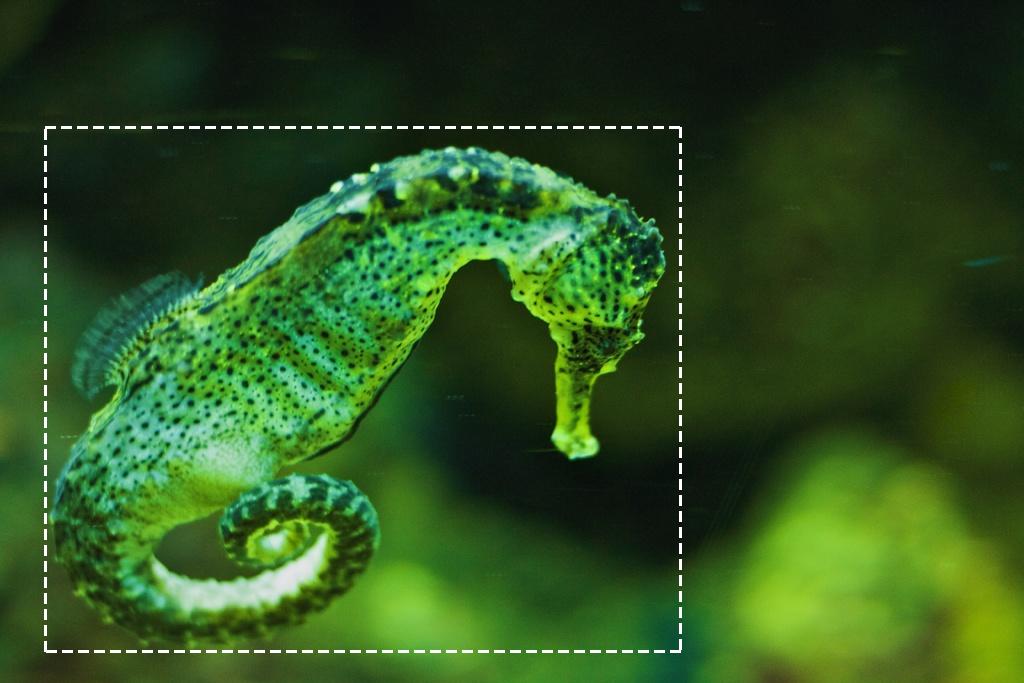}};

\node[inner sep=0, minimum width=1.524cm, minimum height=0.752cm, below = 0.3cm of oi1, anchor = north] (oi2) {\includegraphics[width=0.28\textwidth]{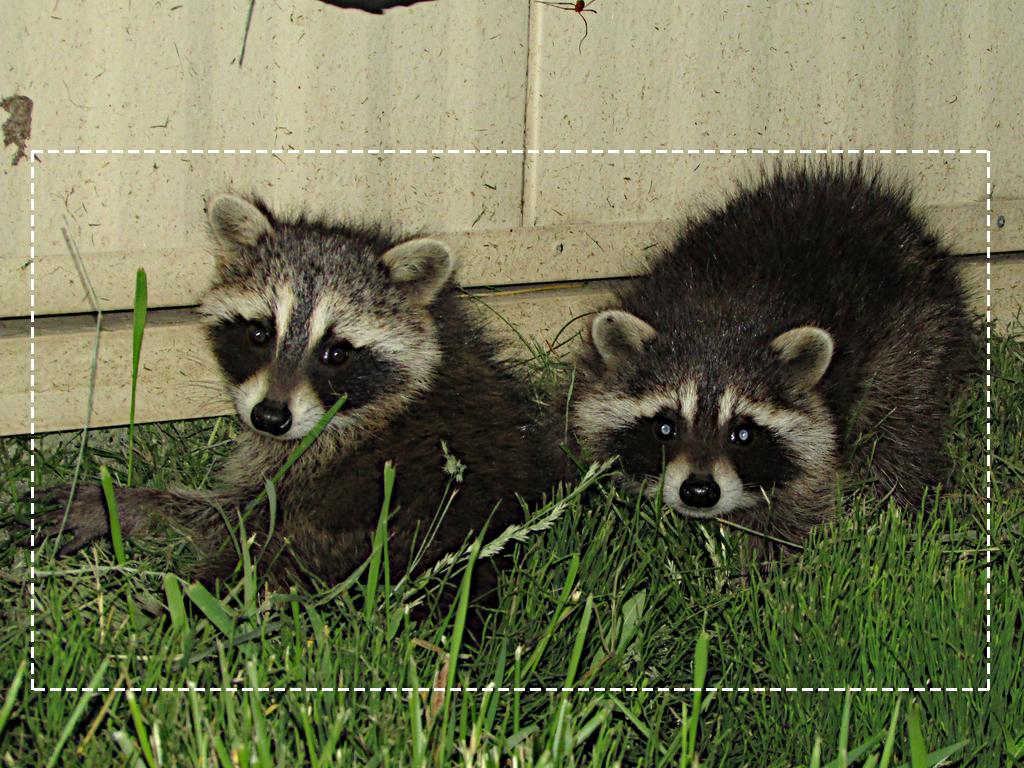}};

\node[inner sep=0, minimum width=1.524cm, minimum height=0.752cm, below = 0.3cm of oi2, anchor = north] (oi3) {\includegraphics[width=0.28\textwidth]{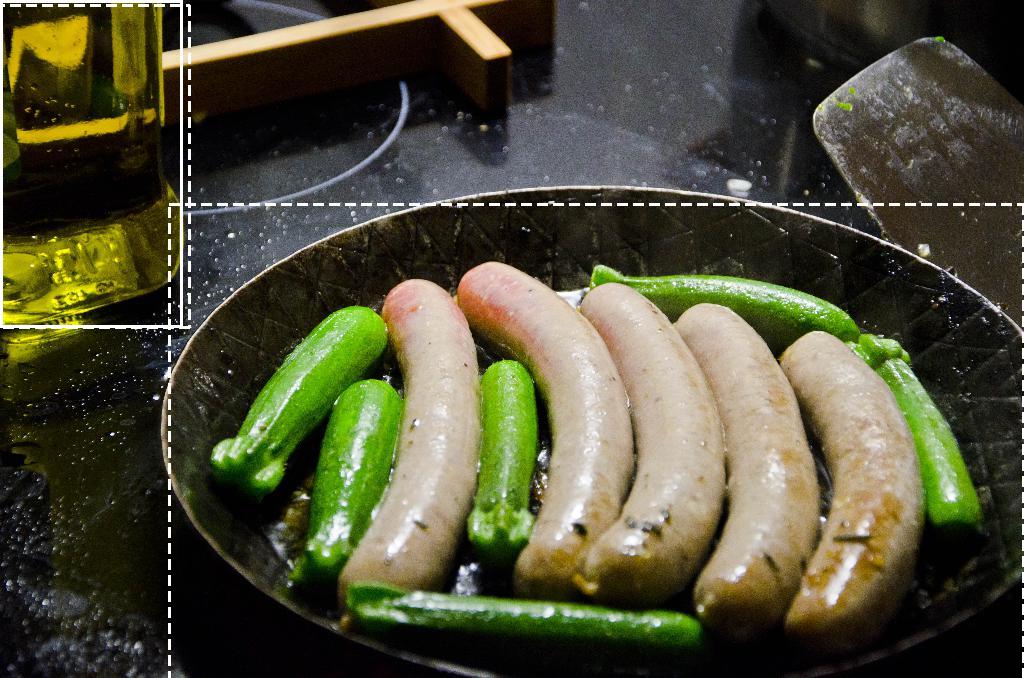}};


\draw[thick] ($(voc1.east) + (0.35, 3.0)$) -- ($(voc1.east) + (0.35, -9.59)$);

\draw[thick] ($(oi2.south) + (2.45cm, -0.15)$) -- ($(oi2.south) + (-13.66cm, -0.15)$);

\draw[thick] ($(oi1.north) + (2.45cm, 0.25)$) -- ($(oi1.north) + (-13.66cm, 0.25)$);


\node [above = 0.5cm of voc1.north, anchor=south, align=center] {ID dataset \\ Pascal VOC};

\node [above = 0.5cm of coco1.north, anchor=south, align=center] {MS-COCO};

\node [above = 0.5cm of oi1.north, anchor=south, align=center] {OpenImages};

\node [above = 2.5cm of coco1.east, anchor=south, align=center, yshift = 0.25cm] {OOD datasets};

\node [left = 0.5cm of voc3.west, anchor=east, align=center, rotate=90, xshift=1cm] {Failure cases};

\node [left = 0.5cm of voc1.west, anchor=east, align=center, rotate=90, xshift=-1cm] {Success cases};

\coordinate (legendNW) at ($(voc3.south west)+(0,-0.2cm)$);
\coordinate (legendSE) at ($(oi3.south east)+(0,-1.3cm)$);

\draw[thick] (legendNW) rectangle (legendSE);

\coordinate (legendLeftCenter) at ($(legendNW)!0.5!(legendNW |- legendSE) + (0.2cm,0)$);

\node[
    rectangle,
    draw,
    dashed,
    minimum width=0.5cm,
    minimum height=0.5cm,
    inner sep=0pt,
    anchor=west
] (legendDashedBox) at (legendLeftCenter) {};

\node[
    right=0.2cm of legendDashedBox,
    anchor=west,
    align=left
]
(od_desc)
{
    \texttt{DeformableDETR} prediction
};

\node[
    rectangle,
    draw,
    minimum width=0.5cm,
    minimum height=0.5cm,
    inner sep=0pt,
    anchor=west,
    right = 0.2cm of od_desc
] (boxsam3) {};

\node[
    right=0.2cm of boxsam3,
    anchor=west,
    align=left
]
(sam3_desc)
{
    \texttt{SAM3} prediction
};

\node[
    right=0.2cm of sam3_desc,
    anchor=west,
    align=left
]
(clas_desc)
{
    Classified as:
};

\node[
    right=0.2cm of clas_desc,
    anchor=west,
    align=left
]
(id_desc)
{
    \textcolor{green}{\cmark} ID
};

\node[
    right=0.2cm of id_desc,
    anchor=west,
    align=left
]
(ood_desc)
{
    \textcolor{tu3}{\xmark} OOD
};

\node[rectangle,fill=white, fill opacity=0.9, below = 0.01 cm of voc1.north, thick, anchor=north, xshift=-0.02cm, text=black, text opacity=1] (bbox1_tag) {\footnotesize{Person} \textcolor{green}{\cmark}};

\node[rectangle,fill=white, fill opacity=0.9, below = 0.55 cm of voc1.north, thick, anchor=north, xshift=-1.26cm, text=black, text opacity=1] (bbox2_tag) {\footnotesize{Horse} \textcolor{green}{\cmark}};

\node[rectangle,fill=white, fill opacity=0.9, below = 0.05 cm of voc2.north, thick, anchor=north, xshift=-1.87cm, text=black, text opacity=1] (bbox3_tag) {\footnotesize{Bird} \textcolor{green}{\cmark}};

\node[rectangle,fill=white, fill opacity=0.9, below = 1.05 cm of voc3.north, thick, anchor=north, xshift=0.43cm, text=black, text opacity=1] (bbox4_tag) {\footnotesize{Chair} \textcolor{tu3}{\xmark}};
\node[rectangle,fill=white, fill opacity=0.9, below = 0.16 cm of coco1.north, thick, anchor=north, xshift=-1.05cm, text=black, text opacity=1] (bbox5_tag) {\footnotesize{Cow} \textcolor{tu3}{\xmark}};

\node[rectangle,fill=white, fill opacity=0.9, below = 0.35 cm of coco1.north, thick, anchor=north, xshift=1.06cm, text=black, text opacity=1] (bbox6_tag) {\footnotesize{Cow} \textcolor{tu3}{\xmark}};

\node[rectangle,fill=white, fill opacity=0.9, below = 0.35 cm of coco2.north, thick, anchor=north, xshift=-1.16cm, text=black, text opacity=1] (bbox7_tag) {\footnotesize{Boat} \textcolor{tu3}{\xmark}};

\node[rectangle,fill=white, fill opacity=0.9, below = 0.00 cm of coco3.north, thick, anchor=north, xshift=-1.45cm, text=black, text opacity=1] (bbox8_tag) {\footnotesize{Potted plant} \textcolor{green}{\cmark}};

\node[rectangle,fill=white, fill opacity=0.9, below = 2.40 cm of coco3.north, thick, anchor=north, xshift=-0.20cm, text=black, text opacity=1] (bbox9_tag) {\footnotesize{Potted plant} \textcolor{tu3}{\xmark}};

\node[rectangle,fill=white, fill opacity=0.9, below = 0.15 cm of oi1.north, thick, anchor=north, xshift=-1.76cm, text=black, text opacity=1] (bbox10_tag) {\footnotesize{Bird} \textcolor{tu3}{\xmark}};

\node[rectangle,fill=white, fill opacity=0.9, below = 0.25 cm of oi2.north, thick, anchor=north, xshift=-1.88cm, text=black, text opacity=1] (bbox11_tag) {\footnotesize{Cat} \textcolor{tu3}{\xmark}};

\node[rectangle,fill=white, fill opacity=0.9, below = 0.03 cm of oi3.north, thick, anchor=north, xshift=-1.9cm, text=black, text opacity=1] (bbox12_tag) {\footnotesize{Bottle} \textcolor{green}{\cmark}};

\node[rectangle,fill=white, fill opacity=0.9, below = 0.48 cm of oi3.north, thick, anchor=north, xshift=1.61cm, text=black, text opacity=1] (bbox13_tag) {\footnotesize{Potted plant} \textcolor{tu3}{\xmark}};

\end{tikzpicture}

%% file: main.bbl
\begin{thebibliography}{50}
\providecommand{\natexlab}[1]{#1}
\providecommand{\url}[1]{\texttt{#1}}
\expandafter\ifx\csname urlstyle\endcsname\relax
  \providecommand{\doi}[1]{doi: #1}\else
  \providecommand{\doi}{doi: \begingroup \urlstyle{rm}\Url}\fi

\bibitem[Bär et~al.(2023)Bär, Uhrig, Umesh, Cordts, and Fingscheidt]{Baer2023a}
Andreas Bär, Jonas Uhrig, Jeethesh~Pai Umesh, Marius Cordts, and Tim Fingscheidt.
\newblock {A Novel Benchmark for Refinement of Noisy Localization Labels in Autolabeled Datasets for Object Detection}.
\newblock In \emph{Proc. of CVPR - Workshops}, pages 3851--3860, Vancouver, BC, Canada, 2023.

\bibitem[Carion et~al.(2025)Carion, Gustafson, Hu, Debnath, Hu, Suris, Ryali, Alwala, Khedr, Huang, Lei, Ma, Guo, Kalla, Marks, Greer, Wang, Sun, Rädle, Afouras, Mavroudi, Xu, Wu, Zhou, Momeni, Hazra, Ding, Vaze, Porcher, Li, Li, Kamath, Cheng, Dollár, Ravi, Saenko, Zhang, and Feichtenhofer]{carion2026sam3segmentconcepts}
Nicolas Carion, Laura Gustafson, Yuan-Ting Hu, Shoubhik Debnath, Ronghang Hu, Didac Suris, Chaitanya Ryali, Kalyan~Vasudev Alwala, Haitham Khedr, Andrew Huang, Jie Lei, Tengyu Ma, Baishan Guo, Arpit Kalla, Markus Marks, Joseph Greer, Meng Wang, Peize Sun, Roman Rädle, Triantafyllos Afouras, Effrosyni Mavroudi, Katherine Xu, Tsung-Han Wu, Yu Zhou, Liliane Momeni, Rishi Hazra, Shuangrui Ding, Sagar Vaze, Francois Porcher, Feng Li, Siyuan Li, Aishwarya Kamath, Ho~Kei Cheng, Piotr Dollár, Nikhila Ravi, Kate Saenko, Pengchuan Zhang, and Christoph Feichtenhofer.
\newblock {SAM 3: Segment Anything with Concepts}.
\newblock \emph{arXiv:2511.16719}, 2025.

\bibitem[Chen et~al.(2023)Chen, Yao, Song, Li, Rao, and Zhang]{chen2023plot}
Guangyi Chen, Weiran Yao, Xiangchen Song, Xinyue Li, Yongming Rao, and Kun Zhang.
\newblock {PLOT: Prompt Learning with Optimal Transport for Vision-Language Models}.
\newblock In \emph{Proc. of ICLR}, pages 1--13, Kigali, Rwanda, 2023.

\bibitem[Ding and Pang(2024)]{ding2024zero}
Choubo Ding and Guansong Pang.
\newblock {Zero-Shot Out-of-Distribution Detection with Outlier Label Exposure}.
\newblock In \emph{Proc. of IJCNN}, pages 1--8, Yokohama, Japan, 2024.

\bibitem[Du et~al.(2022{\natexlab{a}})Du, Gozum, Ming, and Li]{Xuefeng2022b}
Xuefeng Du, Gabriel Gozum, Yifei Ming, and Yixuan Li.
\newblock {SIREN: Shaping Representations for Detecting Out-of-Distribution Objects}.
\newblock In \emph{Proc. of NeurIPS}, pages 20434--20449, virtual, 2022{\natexlab{a}}.

\bibitem[Du et~al.(2022{\natexlab{b}})Du, Wang, Cai, and Li]{Du2022}
Xuefeng Du, Zhaoning Wang, Mu Cai, and Yixuan Li.
\newblock {VOS: Learning What You Don't Know by Virtual Outlier Synthesis}.
\newblock In \emph{Proc. of ICLR}, pages 1--14, virtual, 2022{\natexlab{b}}.

\bibitem[Esmaeilpourcharandabi et~al.(2022)Esmaeilpourcharandabi, Liu, Robertson, and Shu]{Esmaeilpourcharandabi2022}
Sepideh Esmaeilpourcharandabi, Bing Liu, Eric Robertson, and Lei Shu.
\newblock {Zero-Shot Out-of-Distribution Detection Based on the Pre-trained Model CLIP}.
\newblock In \emph{Proc. of AAAI}, pages 6568--6576, virtual, 2022.

\bibitem[Everingham et~al.(2010)Everingham, Van~Gool, Williams, Winn, and Zisserman]{everingham2010pascal}
Mark Everingham, Luc Van~Gool, Christopher K.~I. Williams, John Winn, and Andrew Zisserman.
\newblock {The PASCAL Visual Object Classes (VOC) Challenge}.
\newblock \emph{International Journal of Computer Vision}, 88\penalty0 (2):\penalty0 303--338, 2010.

\bibitem[Fingscheidt et~al.(2022)Fingscheidt, Gottschalk, and Houben]{Fingscheidt2022}
Tim Fingscheidt, Hanno Gottschalk, and Sebastian Houben, editors.
\newblock \emph{{Deep Neural Networks and Data for Automated Driving: Robustness, Uncertainty Quantification, and Insights Towards Safety}}.
\newblock Springer International Publishing, Cham, 2022.

\bibitem[Gao et~al.(2023)Gao, Zhao, Hong, and Xu]{Gao_2023_ICCV}
Ruiyuan Gao, Chenchen Zhao, Lanqing Hong, and Qiang Xu.
\newblock {DIFFGUARD: Semantic Mismatch-Guided Out-of-Distribution Detection Using Pre-Trained Diffusion Models}.
\newblock In \emph{Proc. of ICCV}, pages 1579--1589, Paris, France, 2023.

\bibitem[Graham et~al.(2023)Graham, Pinaya, Tudosiu, Nachev, Ourselin, and Cardoso]{Graham_2023_CVPR}
Mark~S. Graham, Walter~H.L. Pinaya, Petru-Daniel Tudosiu, Parashkev Nachev, Sebastien Ourselin, and Jorge Cardoso.
\newblock {Denoising Diffusion Models for Out-of-Distribution Detection}.
\newblock In \emph{Proc. of CVPR - Workshops}, pages 2948--2957, Vancouver, BC, Canada, 2023.

\bibitem[Hendrycks and Gimpel(2017)]{Hendrycks2017}
Dan Hendrycks and Kevin Gimpel.
\newblock {A Baseline for Detecting Misclassified and Out-of-Distribution Examples in Neural Networks}.
\newblock In \emph{Proc. of ICLR}, pages 1--12, Toulon, France, 2017.

\bibitem[Hendrycks et~al.(2022)Hendrycks, Basart, Mazeika, Zou, Kwon, Mostajabi, Steinhardt, and Song]{pmlr-v162-hendrycks22a}
Dan Hendrycks, Steven Basart, Mantas Mazeika, Andy Zou, Joseph Kwon, Mohammadreza Mostajabi, Jacob Steinhardt, and Dawn Song.
\newblock {Scaling Out-of-Distribution Detection for Real-World Settings}.
\newblock In \emph{Proc. of ICML}, pages 8759--8773, Baltimore, MD, USA, 2022.

\bibitem[Hsu et~al.(2020)Hsu, Shen, Jin, and Kira]{Hsu_2020_CVPR}
Yen-Chang Hsu, Yilin Shen, Hongxia Jin, and Zsolt Kira.
\newblock {Generalized ODIN: Detecting Out-of-Distribution Image Without Learning From Out-of-Distribution Data}.
\newblock In \emph{Proc. of CVPR}, pages 10951--10960, Seattle, WA, USA, 2020.

\bibitem[Jiang et~al.(2024)Jiang, Liu, Fang, Chen, Liu, Zheng, and Han]{jiang2024negative}
Xue Jiang, Feng Liu, Zhen Fang, Hong Chen, Tongliang Liu, Feng Zheng, and Bo Han.
\newblock {Negative Label Guided {OOD} Detection with Pretrained Vision-Language Models}.
\newblock In \emph{Proc. of ICLR}, pages 1--12, Vienna, Austria, 2024.

\bibitem[Kuznetsova et~al.(2020)Kuznetsova, Rom, Alldrin, Uijlings, Krasin, Pont{-}Tuset, Kamali, Popov, Malloci, Duerig, and Ferrari]{Kuznetsova2020}
Alina Kuznetsova, Hassan Rom, Neil Alldrin, Jasper R.~R. Uijlings, Ivan Krasin, Jordi Pont{-}Tuset, Shahab Kamali, Stefan Popov, Matteo Malloci, Tom Duerig, and Vittorio Ferrari.
\newblock {The Open Images Dataset {V4:} Unified Image Classification, Object Detection, and Visual Relationship Detection at Scale}.
\newblock \emph{International Journal of Computer Vision (IJCV)}, 128\penalty0 (7):\penalty0 1956–--1981, 2020.

\bibitem[Lee et~al.(2018)Lee, Lee, Lee, and Shin]{NEURIPS2018_abdeb6f5}
Kimin Lee, Kibok Lee, Honglak Lee, and Jinwoo Shin.
\newblock {A Simple Unified Framework for Detecting Out-of-Distribution Samples and Adversarial Attacks}.
\newblock In \emph{Proc. of NeurIPS}, pages 7167-- 7177, Montr\' {e}al, QC, Canada, 2018.

\bibitem[Li et~al.(2023)Li, Chen, He, Yu, Liu, and Jia]{Li_2023_CVPR}
Jingyao Li, Pengguang Chen, Zexin He, Shaozuo Yu, Shu Liu, and Jiaya Jia.
\newblock {Rethinking Out-of-Distribution (OOD) Detection: Masked Image Modeling Is All You Need}.
\newblock In \emph{Proc. of CVPR}, pages 11578--11589, Vancouver, BC, Canada, 2023.

\bibitem[Li et~al.(2026)Li, Motamedi, Kwon, Park, and Kim]{Li_2026_CVPR}
Yuchuan Li, Azadeh Motamedi, Hyock~Ju Kwon, Chul~B Park, and Il-Min Kim.
\newblock {UNI-OOD: Unified Object- and Image-level Out-of-Distribution Detection via Cross-Context Attentive Vision-Language Modeling}.
\newblock In \emph{Proc.\ of CVPR}, pages 6282--6292, Denver, CO, USA, 2026.

\bibitem[Liang et~al.(2018)Liang, Li, and Srikant]{Liang2018}
S. Liang, Y. Li, and R. Srikant.
\newblock {Enhancing the Reliability of Out-Of-Distribution Image Detection in Neural Networks}.
\newblock In \emph{Proc. of ICLR}, pages 1--27, Vancouver, BC, Canada, 2018.

\bibitem[Lin et~al.(2014)Lin, Maire, Belongie, Hays, Perona, Ramanan, Doll{\'a}r, and Zitnick]{lin2014microsoft}
Tsung-Yi Lin, Michael Maire, Serge Belongie, James Hays, Pietro Perona, Deva Ramanan, Piotr Doll{\'a}r, and C~Lawrence Zitnick.
\newblock {Microsoft COCO: Common Objects in Context}.
\newblock In \emph{Proc. of ECCV}, pages 740--755, Zurich, Switzerland, 2014.

\bibitem[Liu et~al.(2020)Liu, Wang, Owens, and Li]{liu2020energy}
Weitang Liu, Xiaoyun Wang, John Owens, and Yixuan Li.
\newblock {Energy-based Out-of-Distribution Detection}.
\newblock In \emph{Proc. of NeurIPS}, pages 21464--21475, virtual, 2020.

\bibitem[Liu and Zach(2024)]{Liu2024}
Xixi Liu and Christopher Zach.
\newblock {TAG: Text Prompt Augmentation for Zero-Shot Out-of-Distribution Detection´}.
\newblock In \emph{Proc. of ECCV}, pages 364--380, Milano, Italy, 2024.

\bibitem[Liu et~al.(2023)Liu, Zhou, Wang, and Weinberger]{liu2023}
Zhenzhen Liu, Jin~Peng Zhou, Yufan Wang, and Kilian~Q Weinberger.
\newblock {Unsupervised Out-of-Distribution Detection with Diffusion Inpainting}.
\newblock In \emph{Proc. of ICML}, pages 22528--22538, Honolulu, HI, USA, 2023.

\bibitem[Lu et~al.(2024)Lu, Gong, Wang, Xue, Yao, and Moore]{lu2024learning}
Haodong Lu, Dong Gong, Shuo Wang, Jason Xue, Lina Yao, and Kristen Moore.
\newblock {Learning with Mixture of Prototypes for Out-of-Distribution Detection}.
\newblock In \emph{Proc. of ICLR}, pages 1--14, Vienna, Austria, 2024.

\bibitem[Ming et~al.(2022)Ming, Cai, Gu, Sun, Li, and Li]{Ming2022}
Yifei Ming, Ziyang Cai, Jiuxiang Gu, Yiyou Sun, Wei Li, and Yixuan Li.
\newblock {Delving into Out-of-Distribution Detection with Vision-Language Representations}.
\newblock In \emph{Proc. of NeurIPS}, pages 35087--35102, virtual, 2022.

\bibitem[Ming et~al.(2023)Ming, Sun, Dia, and Li]{ming2023how}
Yifei Ming, Yiyou Sun, Ousmane Dia, and Yixuan Li.
\newblock {How to Exploit Hyperspherical Embeddings for Out-of-Distribution Detection?}
\newblock In \emph{Proc. of ICLR}, pages 1--13, Kigali, Rwanda, 2023.

\bibitem[Miyai et~al.(2023)Miyai, Yu, Irie, and Aizawa]{NEURIPS2023_f0606b88}
Atsuyuki Miyai, Qing Yu, Go Irie, and Kiyoharu Aizawa.
\newblock {LoCoOp: Few-Shot Out-of-Distribution Detection via Prompt Learning}.
\newblock In \emph{Proc. of NeurIPS}, pages 76298--76310, New Orleans, LA, USA, 2023.

\bibitem[Miyai et~al.(2025)Miyai, Yu, Irie, and Aizawa]{Miyai2025}
Atsuyuki Miyai, Qing Yu, Go Irie, and Kiyoharu Aizawa.
\newblock {GL-MCM: Global and Local Maximum Concept Matching for Zero-Shot Out-of-Distribution Detection}.
\newblock \emph{International Journal of Computer Vision}, 133\penalty0 (6):\penalty0 3586–--3596, 2025.

\bibitem[Nguyen et~al.(2026)Nguyen, Zhou, Liu, Bui, Weinberger, Chao, and Le]{Nguyen2026}
Quang-Huy Nguyen, Jin~Peng Zhou, Zhenzhen Liu, Khanh-Huyen Bui, Kilian~Q. Weinberger, Wei-Lun Chao, and Dung~D. Le.
\newblock {Detecting Out-of-Distribution Objects through Class-Conditioned Inpainting}.
\newblock In \emph{Proc. of WACV}, pages 1937--1947, Tucson, AZ, USA, 2026.

\bibitem[Radford et~al.(2021)Radford, Kim, Hallacy, Ramesh, Goh, Agarwal, Sastry, Askell, Mishkin, Clark, Krueger, and Sutskever]{radford2021learningtransferablevisualmodels}
Alec Radford, Jong~Wook Kim, Chris Hallacy, Aditya Ramesh, Gabriel Goh, Sandhini Agarwal, Girish Sastry, Amanda Askell, Pamela Mishkin, Jack Clark, Gretchen Krueger, and Ilya Sutskever.
\newblock {Learning Transferable Visual Models From Natural Language Supervision}.
\newblock In \emph{Proc. of ICML}, pages 8748--8763, virtual, 2021.

\bibitem[Ren et~al.(2019)Ren, Liu, Fertig, Snoek, Poplin, Depristo, Dillon, and Lakshminarayanan]{Jie2019}
Jie Ren, Peter~J. Liu, Emily Fertig, Jasper Snoek, Ryan Poplin, Mark Depristo, Joshua Dillon, and Balaji Lakshminarayanan.
\newblock {Likelihood Ratios for Out-of-Distribution Detection}.
\newblock In \emph{Proc. of NeurIPS}, pages 14707--14718, Vancouver, BC, Canada, 2019.

\bibitem[Saijo et~al.(2025)Saijo, Zhang, Scheibler, Cornell, Li, Ni, Kumar, Sach, Fu, Fingscheidt, and Watanabe]{Saijo2025}
Kohei Saijo, Wangyou Zhang, Robin Scheibler, Samuele Cornell, Chenda Li, Zhaoheng Ni, Anurag Kumar, Marvin Sach, Yihui Fu, Tim Fingscheidt, and Shinji Watanabe.
\newblock {Interspeech 2025 URGENT Speech Enhancement Challenge}.
\newblock In \emph{Proc.\ of Interspeech}, pages 858--862, Rotterdam, The Netherlands, 2025.

\bibitem[Sastry and Oore(2020)]{sastry2020}
Chandramouli~Shama Sastry and Sageev Oore.
\newblock {Detecting Out-of-Distribution Examples with {G}ram Matrices}.
\newblock In \emph{Proc. of ICML}, pages 8491--8501, virtual, 2020.

\bibitem[Schlegl et~al.(2017)Schlegl, Seeb{\"o}ck, Waldstein, Schmidt-Erfurth, and Langs]{10.1007/978-3-319-59050-9_12}
Thomas Schlegl, Philipp Seeb{\"o}ck, Sebastian~M. Waldstein, Ursula Schmidt-Erfurth, and Georg Langs.
\newblock Unsupervised anomaly detection with generative adversarial networks to guide marker discovery.
\newblock In \emph{Proc. of Information Processing in Medical Imaging}, pages 146--157, Cham, 2017. Springer International Publishing.

\bibitem[Sun et~al.(2022)Sun, Ming, Zhu, and Li]{sun2022}
Yiyou Sun, Yifei Ming, Xiaojin Zhu, and Yixuan Li.
\newblock {Out-of-Distribution Detection with Deep Nearest Neighbors}.
\newblock In \emph{Proc. of ICML}, pages 20827--20840, Baltimore, MD, USA, 2022.

\bibitem[Tack et~al.(2020)Tack, Mo, Jeong, and Shin]{Tack2020}
Jihoon Tack, Sangwoo Mo, Jongheon Jeong, and Jinwoo Shin.
\newblock {CSI: Novelty Detection via Contrastive Learning on Distributionally Shifted Instances}.
\newblock In \emph{Proc. of NeurIPS}, pages 11839--11852, virtual, 2020.

\bibitem[Wang et~al.(2023)Wang, Li, Yao, and Li]{Wang_2023_ICCV}
Hualiang Wang, Yi Li, Huifeng Yao, and Xiaomeng Li.
\newblock {CLIPN for Zero-Shot OOD Detection: Teaching CLIP to Say No}.
\newblock In \emph{Proc. of ICCV}, pages 1802--1812, Paris, France, 2023.

\bibitem[Wei et~al.(2022)Wei, Xie, Cheng, Feng, An, and Li]{pmlr-v162-wei22d}
Hongxin Wei, Renchunzi Xie, Hao Cheng, Lei Feng, Bo An, and Yixuan Li.
\newblock {Mitigating Neural Network Overconfidence with Logit Normalization}.
\newblock In \emph{Proc. of ICML}, pages 23631--23644, Baltimore, MD, USA, 2022.

\bibitem[Wilson et~al.(2023)Wilson, Fischer, Dayoub, Miller, and Sunderhauf]{Wilson2023}
Samuel Wilson, Tobias Fischer, Feras Dayoub, Dimity Miller, and Niko Sunderhauf.
\newblock { SAFE: Sensitivity-Aware Features for Out-of-Distribution Object Detection}.
\newblock In \emph{Proc. of ICCV}, pages 23508--23519, Los Alamitos, CA, USA, 2023.

\bibitem[Wu and Deng(2023)]{DBLP:conf/cvpr/WuD23}
Aming Wu and Cheng Deng.
\newblock {Discriminating Known from Unknown Objects via Structure-Enhanced Recurrent Variational AutoEncoder}.
\newblock In \emph{Proc. of CVPR}, pages 23956--23965, Vancouver, BC, Canada, 2023.

\bibitem[Wu and Deng(2025)]{11093101}
Aming Wu and Cheng Deng.
\newblock {Percept, Memory, and Imagine: World Feature Simulating for Open-Domain Unknown Object Detection}.
\newblock In \emph{Proc. of CVPR}, pages 4682--4691, Nashville, TN, USA, 2025.

\bibitem[Wu et~al.(2023)Wu, Chen, and Deng]{Wu2023_Iccv}
Aming Wu, Da Chen, and Cheng Deng.
\newblock {Deep Feature Deblurring Diffusion for Detecting Out-of-Distribution Objects}.
\newblock In \emph{Proc. of ICCV}, pages 13335--13345, Paris, France, 2023.

\bibitem[Xiao et~al.(2020)Xiao, Yan, and Amit]{Xiao2020}
Zhisheng Xiao, Qing Yan, and Yali Amit.
\newblock {Likelihood Regret: An Out-of-Distribution Detection Score For Variational Auto-encoder}.
\newblock In \emph{Proc. of NeurIPS}, pages 20685--20696, virtual, 2020.

\bibitem[Xiao et~al.(2021)Xiao, Yan, and Amit]{xiao2021reallyneedlearnrepresentations}
Zhisheng Xiao, Qing Yan, and Yali Amit.
\newblock {Do We Really Need to Learn Representations from In-domain Data for Outlier Detection?}
\newblock \emph{arXiv:2105.09270}, 2021.

\bibitem[Yu et~al.(2018)Yu, Xian, Chen, Liu, Liao, Madhavan, and Darrell]{Yu2018b}
Fisher Yu, Wenqi Xian, Yingying Chen, Fangchen Liu, Mike Liao, Vashisht Madhavan, and Trevor Darrell.
\newblock {BDD100K: A Diverse Driving Video Database With Scalable Annotation Tooling}.
\newblock \emph{arXiv:1805.04687}, 2018.

\bibitem[Zhang et~al.(2025)Zhang, Chen, Qu, Li, Lu, Wan, Xiao, and Wang]{zhang2025runaobjectleveloutofdistributiondetection}
Bin Zhang, Jinggang Chen, Xiaoyang Qu, Guokuan Li, Kai Lu, Jiguang Wan, Jing Xiao, and Jianzong Wang.
\newblock {RUNA: Object-level Out-of-Distribution Detection via Regional Uncertainty Alignment of Multimodal Representations}.
\newblock In \emph{Proc. of AAAI}, pages 26418--26426, Philadelphia, PA, USA, 2025.

\bibitem[Zhou et~al.(2022)Zhou, Yang, Loy, and Liu]{Zhou2022CoOp}
Kaiyang Zhou, Jingkang Yang, Chen~Change Loy, and Ziwei Liu.
\newblock {Learning to Prompt for Vision-Language Models}.
\newblock \emph{International Journal of Computer Vision}, 130\penalty0 (9):\penalty0 2337--2348, 2022.

\bibitem[Zhu et~al.(2021)Zhu, Su, Lu, Li, Wang, and Dai]{Zhu2021}
Xizhou Zhu, Weijie Su, Lewei Lu, Bin Li, Xiaogang Wang, and Jifeng Dai.
\newblock {Deformable DETR: Deformable Transformers for End-to-End Object Detection}.
\newblock In \emph{Proc. of ICLR}, pages 1--16, virtual, 2021.

\bibitem[Zong et~al.(2018)Zong, Song, Min, Cheng, Lumezanu, Cho, and Chen]{zong2018deep}
Bo Zong, Qi Song, Martin~Renqiang Min, Wei Cheng, Cristian Lumezanu, Daeki Cho, and Haifeng Chen.
\newblock {Deep Autoencoding Gaussian Mixture Model for Unsupervised Anomaly Detection}.
\newblock In \emph{Proc. of ICLR}, pages 1--14, Vancouver, BC, Canada, 2018.

\end{thebibliography}
